\documentclass[12pt,letterpaper]{article}
\usepackage[T1]{fontenc}
\usepackage[utf8]{inputenc}
\usepackage{lmodern}
\usepackage[margin=1in]{geometry}
\usepackage{amsmath,amssymb,amsthm}
\usepackage{algorithm,algpseudocode}
\usepackage{multirow,booktabs,array,longtable}
\usepackage{graphicx}
\usepackage[dvipsnames,svgnames,x11names]{xcolor}
\usepackage{tikz}
\usetikzlibrary{fit}
\usepackage{caption,subcaption}
\usepackage{parskip,microtype}
\usepackage{natbib}
\usepackage[hyphens]{url}
\usepackage{xurl}
\usepackage[unicode]{hyperref}
\usepackage{bookmark}
\hypersetup{
  pdftitle={Federated Community Detection in Bipartite Networks},
  pdfauthor={Wanjie Wang},
  pdfkeywords={Spectral clustering; Federated learning; Gradient descent; Power iteration},
  colorlinks=true,
  linkcolor=blue,
  citecolor=blue,
  urlcolor=blue,
  pdfcreator={LaTeX}}
\newtheorem{theorem}{Theorem}

\newtheorem{assumption}{Assumption}

\theoremstyle{plain}
\newtheorem{lemma}{Lemma}

\newtheorem{proposition}{Proposition}
\newtheorem{remark}{Remark}
\theoremstyle{remark}
\newcommand{\gdpi}{\text{FLICKER} }

\def\reals{\mathbb{R}}
\def\be{\begin{equation}}
\def\ee{\end{equation}}
\def\ben{\begin{equation*}}
\def\een{\begin{equation*}}

\def\Prob{{\mathbb P}}
\def\E{{\mathbb E}}

\def\Uhat{{\hat{\xi}}}

\def\Uhat{\widehat{U}}

\def\ic{^\top}

\newcommand{\twoninf}[1]{\left\|#1\right\|_{2,\infty}}

\def\blue{\textcolor{black}}

\title{\bfseries Federated Community Detection in Bipartite Networks}
\author{Wanjie Wang\thanks{Email: \href{mailto:wanjie.wang@nus.edu.sg}{wanjie.wang@nus.edu.sg}.}\\[0.4em]
\small Department of Statistics and Data Science\\
\small Department of Mathematics\\
\small National University of Singapore\\
\small Block S16, Level 7, 6 Science Drive 2\\
\small Singapore 117546}
\date{}
\begin{document}
\maketitle
\begin{abstract}
We study federated spectral estimation when a large matrix is horizontally partitioned across servers that cannot share raw data. We propose FLICKER, a target-specific gradient-based power iteration that recovers a singular direction associated with a prescribed spectral target using only matrix--vector products on local blocks. Different target directions can be recovered independently and in parallel. We establish non-asymptotic convergence guarantees and provide a recursive grid-search procedure for spectral target selection.

Building on FLICKER, we develop BiFLICKER for federated community detection in bipartite networks, together with a sequential projected variant for nearby or repeated singular directions. Under a sparse bipartite stochastic block model, we show that properly rescaled leading singular values from a reference server provide valid global targets, and that Sequential BiFLICKER achieves a vanishing misclustering rate and exact recovery under stronger sparsity conditions. Simulations show that BiFLICKER remains close to centralized spectral clustering and is more robust than existing federated spectral baselines when local sample sizes are small. On a LastFM user--artist network naturally partitioned by country, Sequential BiFLICKER attains an overall ARI of $0.96$ against centralized clustering while revealing cross-country communities with distinct listening profiles.
\end{abstract}
\clearpage

\section{Introduction}
Spectral methods are fundamental to modern statistics and data science, providing a powerful tool for uncovering low-dimensional structures from complex data \citep{von2007tutorial}.
{\color{black}In network community detection, spectral clustering exploits the leading eigenvectors or singular vectors of an adjacency matrix to construct low-dimensional representations of nodes, which are used to recover community memberships \citep{hu2024network, SpecMultiNet, shen2025optimal, shen2026multilayer}. Therefore, singular-vector estimation is a key computational step in spectral community detection. This step is closely related to principal component analysis (PCA), which also relies on leading eigenvectors or singular vectors, although the two methods serve different statistical objectives \citep{denoising, ma2026iIFLearn}.}

{\color{black}Spectral} methods require the critical eigenvectors and eigenvalues of the data matrix $A$. Classical methods, including power iteration \citep{golub2013matrix}, Arnoldi iteration \citep{arnoldi1951principle}, and the Lanczos algorithm \citep{lanczos1950iteration}, are computationally efficient when the full data matrix $A$ is directly accessible. However, modern data are increasingly generated across and stored across multiple institutions, mobile devices, or healthcare systems. In such scenarios, the data is represented by local matrices $A_k$ on each server $k$, making traditional centralized computation impossible.

This shift toward decentralized data, coupled with growing privacy and computation constraints, has necessitated the rise of Federated Learning. Federated learning enables multiple data holders to collaboratively train global models while keeping all raw data local \citep{mcmahan2017communication}, and it has yielded fruitful results in supervised learning. However, federated spectral analysis is still maturing. Early foundations in distributed PCA focused on one-shot communication or sketching techniques to aggregate local covariance matrices \citep{disPCA, garber2017communication, DistPCA}. These often suffer from data heterogeneity and limited sample sizes in local servers. More recent advances utilize iterative strategies to find the global eigenvectors without local eigenvectors. For instance, \cite{li2021communication} proposes to use local power iterations followed by a global synchronization. The Riemannian gradient descent approach in \cite{oja1982simplified} motivates another line of iterative methods, such as \cite{huang2020communication} on the first eigenvector and \cite{FastPCA} without the centre server. Despite these strides, the specific application of these principles to federated spectral clustering, especially for social network data, remains sparse.

\subsection{Federated Learning in Bipartite Networks}\label{sec:intro-bipartite}
The need for federated spectral methods is particularly acute in bipartite networks. A  bipartite network records interactions between two distinct groups of entities, such as users and movies in recommendation systems, consumers and products in e-commerce, or listeners and artists on music platforms. We denote a bipartite network by \(G = (\mathcal{L}, \mathcal{R}, \mathcal{E})\), where \(\mathcal{L}\) and \(\mathcal{R}\) represent two node sets, and \(\mathcal{E} \subseteq \mathcal{L} \times \mathcal{R}\) represents the set of edges or interactions between nodes.
The adjacency matrix $B \in \{0, 1\}^{|\mathcal{L}| \times |\mathcal{R}|}$ summarizes the information in $G$, where $B(i,j) = 1$ if and only if $(i,j) \in \mathcal{E}$.

The community detection problem is of great interest in network analysis.
We want to partition the node set $\mathcal{L}$ into several disjoint communities/clusters \(\mathcal{L}_i \subseteq \mathcal{L}\) and partition $\mathcal{R}$ into \(\mathcal{R}_j \subseteq \mathcal{R}\), such that {all node pairs in $\mathcal{L}_i \times \mathcal{R}_j$ share the same connection probability. }
{Such similarity can arise from shared latent profiles. For example, in a user-artist network, users in the same community may have a common preference profile across artist groups, while artists in the same community may attract a similar mixture of user groups. Recovering these profiles provides a better understanding of user behavior.}
Spectral community detection methods that rely on the left and right singular vectors of $B$ play an important role in recovering this structure \citep{specbi, disim, Dscore}.

We are interested in the federated setting of bipartite networks. Each local server $k$ has observed a partial network $G^{(k)} = (\mathcal{L}^{(k)}, \mathcal{R}^{(k)}, \mathcal{E}^{(k)})$, $1 \leq k \leq K$.
In many settings, all servers share a common node set $\mathcal{R}^{(k)} = \mathcal{R}$, while their left node set $\mathcal{L}^{(k)}$ are disjoint.
For example, country-specific or regional servers may store interactions from different user populations with a common catalog of items or artists.
The challenge is to collaboratively learn global community structures without directly pooling the local interaction matrices.

Motivated by this gap, our work addresses the challenge of detecting the community structures of both $\mathcal{L}$ and $\mathcal{R}$ under federated data with communication constraints.

\subsection{Our Contribution and Organization}
The primary goal of this work is to bridge the gap between decentralized data storage and robust spectral analysis for bipartite networks. Our contributions are as follows.
\begin{itemize}
    \item We propose a gradient-descent-based power iteration that recovers an eigendirection associated with the eigenvalue nearest to the input tuning parameter $\lambda$. It only requires $Av^t$ and $A^2v^t$ in each update, where $v^t$ is the current eigendirection. We establish its non-asymptotic convergence and describe the invariant-subspace case when the nearest eigenvalue is not unique.

    \item We propose the Federated Learning of Iterative Communal Kernels for Eigenvector Recovery (FLICKER) algorithm based on our new power iteration method. Each local server returns matrix-vector products rather than the full data matrix.
    We provide both left- and right-singular-vector implementations and analyze their communication and computation costs.

    \item We extend this framework to bipartite networks through BiFLICKER. The target-specific iterations can be run in parallel when the target singular values identify distinct directions. {We also introduce Sequential BiFLICKER as a variant, which projects out previously recovered directions and is more robust to nearby targets at the cost of losing parallelism across directions.}

    \item {We establish the community misclustering error bound and exact recovery of BiFLICKER under a sparse bipartite stochastic block model. We demonstrate the method in simulations and apply Sequential BiFLICKER to the LastFM user–artist network, using countries to define the local servers.
    }
\end{itemize}

The remainder of this paper is organized as follows. Section \ref{sec:eigen} details our new gradient descent approach for eigenvector recovery and introduces the FLICKER algorithm. Section \ref{sec:fedcd} discusses spectral community detection for bipartite networks and proposes the BiFLICKER algorithm. Theoretical results regarding the convergence and consistency of these methods are presented in Section~\ref{sec:theory}. Finally, Sections \ref{sec:simu} and \ref{sec:data} evaluate the performance of our approach through numerical studies and real-world data applications.

\section{Federated Learning of Eigenvectors}\label{sec:eigen}

\subsection{Eigenvectors: A New Gradient Descent Approach}
We first re-envision the classical eigenvector recovery problem through the lens of constrained optimization.
The reformulation is motivated by the following lemma.
\begin{lemma}
\label{lem:opt}
Suppose a symmetric matrix $A$ has eigenvalues $\lambda_1>\cdots>\lambda_n$ with corresponding eigenvectors $u_i$, $1 \leq i \leq n$. For any target value $\lambda$, consider the constrained optimization:
\begin{equation}
\label{eqn:P1}
\min_{v: \|v\| = 1} \ell(v; \lambda), \quad \mbox{where } \ell(v; \lambda) = \frac{1}{2}\|\lambda v - A v \|^2.
\end{equation}
The loss function $\ell(v;\lambda)$ has at least $2n$ stationary points at $v = \pm u_i$. Furthermore, the global minimum is achieved at $\hat{v} \in \mbox{span}\{\pm u_{i^*}\}$, $i^*{\in}\arg\min_{1 \leq i \leq n}|\lambda - \lambda_i|$.
\end{lemma}

Lemma \ref{lem:opt} introduces an interesting fact: the global minimizer of $\ell(v;\lambda)$ is achieved at the eigenvector $u_{i^*}$, where $\lambda_{i^*}$ is the eigenvalue nearest to the target value $\lambda$. For example, when we take a very large $\lambda \geq \|A\|$, then the optimizer is the first eigenvector $u_1$. It allows us to recover any eigenvector with a proper target value input $\lambda$.

We present a gradient descent power iteration method based on the insight of Lemma \ref{lem:opt}. For the loss function $\ell(v;\lambda)$, the gradient is
\[
\nabla \ell(v; \lambda) = (\lambda I-A)^2v = \lambda^2 v - 2\lambda Av + A(Av),
\]
and then projected onto the sphere $S^{n-1}$. The detailed algorithm can be formulated as Algorithm \ref{alg:GD}.
In each iteration, we only need to find a matrix-vector product $Av^t$, at a computation cost of $O(n^2)$.
{We further present a variant of the original gradient descent algorithm that constrains the eigenvector to lie in a known subspace defined by $P$.
}
\begin{algorithm}[!ht]
\small
\caption{Gradient descent power iteration}\label{alg:GD}
\begin{algorithmic}
\Require $A$, target $\lambda$, iterations $T$, step size $\eta$; {optionally, projection $P$}.
\Ensure $v^T$, an eigenvector {in the subspace given by $P$} with eigenvalue closest to $\lambda$.
\State Initialize $v^0 \sim \mathcal{N}(0,I_n)$ and normalize it to have unit norm.
\For{$t=0,\ldots,T-1$}
    \State $v^{t+1}=v^t-\eta\{\lambda^2v^t-2\lambda Av^t+A(Av^t)\}$.
    \State $v^{t+1}\gets v^{t+1}/\|v^{t+1}\|$;
    {if $P$ is given, instead set $v^{t+1}\gets Pv^{t+1}/\|Pv^{t+1}\|$.}
\EndFor
\end{algorithmic}
\end{algorithm}

The convergence analysis of Algorithm \ref{alg:GD} cannot be solved by standard convex optimization analysis. The challenge comes from two parts: $\ell(v;\lambda)$ on the sphere $S^{n-1}$ is not convex, and the global minimizer may not be unique, i.e., $\lambda_i$ may not be simple.
We provide the convergence results for simple $\lambda_i$'s in Theorem \ref{thm:GD}. {For eigenvalues with multiplicity greater than 1, the convergence results are in Theorem \ref{thm:GDproj} on the projected version of Algorithm \ref{alg:GD}.}
{Discussion on the choice of $\lambda$ is left to Section \ref{sec:lambda}. }
\begin{theorem}
\label{thm:GD}
Let $v^t$ be the estimated eigenvector in the $t$-th iteration of Algorithm \ref{alg:GD} with target value $\lambda$ and {no projection $P$}.
Let $u_i$ be the $i$-th eigenvector of $A$ and $a_{t, i}=\langle v^t, u_i\rangle$ be the inner product of $v^t$ and $u_i$, then $v^t = \sum_{i = 1}^n a_{t, i} u_i$.
{Suppose $i^*=\arg\min_{1 \leq i\leq n}|\lambda - \lambda_i|$ is unique.}
With a step size $\eta > 0$, for iteration $t$, there is
\[
\sum_{i=1}^n a_{t, i}^2=1,\quad \frac{a_{t, i}}{a_{t, j}}=
\biggl(\frac{1-\eta(\lambda-\lambda_i)^2 }{1-\eta(\lambda-\lambda_j)^2  }\biggr)^t \frac{a_{0, i}}{a_{0, j}}.
\]
Furthermore, for a stepsize $\eta<\frac{1}{\max_{1 \leq i \leq n} |\lambda-\lambda_i|^2}$, define the ratio $r(\eta,\lambda,\lambda_{[n]})
=
\frac{
1-\eta\min_{i\neq i^*}(\lambda-\lambda_i)^2
}{
1-\eta(\lambda-\lambda_{i^*})^2
}
<1$, then for $a_{0,i^*} \neq 0$, iteration $t$,
\[
1 - a_{t, i^*}^2 \leq  \frac{1-a_{0, i^*}^2}{a_{0, i^*}^2}  \{r(\eta,\lambda,\lambda_{[n]})\}^{{2}t}.
\]
{Under the normalized Gaussian initialization in Algorithm~\ref{alg:GD}, for any $\delta\in(0,1)$, with probability at least $1-\delta-\exp(-cn)$,
$|a_{0,i^*}|\ge \frac{c_0\delta}{\sqrt n}$ for some universal constants $c,c_0>0$. Therefore, with the same probability, the eigenvector direction $u_{i^*}$ recovery follows that
\[
\min\{\|v^T-u_{i^*}\|, \|v^T+u_{i^*}\|\}
\le \epsilon
\]
whenever
$T \ge \frac{
\log\!\left(C\sqrt n/(\delta\epsilon)\right)
}{-\log r(\eta,\lambda,\lambda_{[n]})
} = O(\frac{\log (\epsilon/n)}{\log r})$ iterations.}
\end{theorem}

Theorem \ref{thm:GD} establishes the convergence rate and hence the computational cost.
\blue{For a fixed matrix $A$, target $\lambda$, and step size
$\eta$, the ratio $r<1$ is constant with respect to the
iteration number, yielding geometric convergence.
To investigate the asymptotic computational complexity, we let $n \to \infty$ and consider a sequence of matrices $A_n$, and therefore the corresponding ratio $r_n$ may depend on $n$.
To achieve an accuracy $\epsilon$, Theorem~\ref{thm:GD} indicates that we need $T=O(\log(\sqrt{n}/(\delta\epsilon))/(-\log r_n))$ iterations.
If $r_n$ is uniformly bounded away from 1, this reduces
to $T=O(\log n)$ for fixed $\epsilon$ and $\delta$,
yielding an overall computational cost of
$O(n^2\log n)$ for dense matrices.}

Theorem \ref{thm:GD} also guides the choice of the step size $\eta$. When the step size $\eta$ satisfies that $\eta < 1/(\max_i |\lambda-\lambda_i|)^2$, the concentration rate $r$ decreases with $\eta$.
Therefore, a proper step size $\eta$ should be as large as $\eta < 1/(\max_i |\lambda-\lambda_i|)^2$ holds.
For example, if $0 < \lambda_i < 1$ holds for all $i$, then $\eta$ can be chosen as $\eta =\max\{\lambda, 1 - \lambda\}^{-2}$ for $\lambda > 0$.

As a comparison, a popular method to recover eigenvectors is the inverse power method \citep{golub2013matrix}, where $v^t = (\lambda I-A)^{-t} v^{0}$ in the $t$-th iteration.
The updates require $(\lambda I-A)^{-1}$, which can be $O(n^3)$ in general. The Krylov approximation method \citep{saad2011numerical} approximates $(\lambda I-A)^{-1}v^t$ on the linear space of $\{v^t, A v^t,\ldots, A^J v^t\}$, which reduces the computational cost to $O(n^2)$ per iteration.
Our method is based only on the three vectors $\{v^t, A v^t, A^2 v^t\}$, which is much simpler. More importantly, our method can be easily adapted to the federated learning scenario where only partial $A$ is known.

{Theorem \ref{thm:GD} focuses on the case that the nearest eigenvalue $\lambda_{i^\star}$ is simple.
Consider a target value $\lambda$ where the nearest eigenvalue $\lambda_{i^\star}$ has multiplicity $s \ge 2$, then Algorithm~\ref{alg:GD} without projection converges to a unit vector in the invariant subspace $\mathcal U_\star := \mathrm{span}\{u_j:\ \lambda_j=\lambda_{i^\star}\}$. }

{To estimate full subspace, we sequentially apply Algorithm~\ref{alg:GD} with the same target value $\lambda$, with a projection $P=I-\sum_k \hat{u}_k \hat{u}_k^\top$ for $\hat{u}_k$ being the previously recovered eigenvectors of $A$.
In detail, we consider an orthonormal eigenbasis $\{u_i\}_{i=1}^n$ of $A$ and a subset $I\subset[n]$ as the eigendirections to be projected out.
For the remaining subspace, let $J_\star=\{j\notin I: |\lambda-\lambda_j|=\min_{\ell\notin I}|\lambda-\lambda_\ell|\}$, the corresponding subspace $\mathcal U_\star=\mathrm{span}\{u_j:j\in J_\star\}$, and the projection $P_\star=\sum_{j\in J_\star}u_j u_j^\top$.
Theorem \ref{thm:GDproj} demonstrates that the recovered $v^T$ almost lies in $U_\star$.}
{
\begin{theorem}
\label{thm:GDproj}
Consider an orthonormal eigenbasis $\{u_i\}_{i=1}^n$ of $A$ and a subset $I \subset [n]$.
Define $P_I^\perp=I_n-\sum_{i\in I}u_i u_i^\top$.
Let $v^t$ be the $t$-th iteration output of Algorithm~\ref{alg:GD} with an orthogonal projection $P$ satisfying $\|P-P_I^\perp\|\le\epsilon_0$.

Suppose that $0<\eta<1/\max_i|\lambda-\lambda_i|^2$.
Define $r_I=\frac{\max_{j\notin I\cup J_\star}[1-\eta(\lambda-\lambda_j)^2]}{\max_{j\notin I}[1-\eta(\lambda-\lambda_j)^2]}<1$, and set it to be 0 if $I\cup J_\star = [n]$.
Then, for any fixed $\varsigma\in(0,1)$, there exists a constant $C_\varsigma>0$ such that, with probability at least $1-\varsigma$
over a normalized Gaussian initialization $v^0$ independent of $P$,
\[
\operatorname{dist}(v^T,\mathcal U_\star)
\le
\epsilon_1+
C_\varsigma\frac{\sqrt n\,T}{\max_{j\notin I}[1-\eta(\lambda-\lambda_j)^2]}\epsilon_0,
\]
where
$\operatorname{dist}(v,\mathcal U_\star)=\|(I_n-P_\star)v\|$ and $T\ge C_\varsigma\log(\epsilon_1/n)/\log r_I$.
\end{theorem}
}

{Theorem \ref{thm:GDproj} indicates that the recovery error will depend on $\epsilon_0$, the error in previously found $v_k$. To mitigate this, one may use a stricter accuracy threshold $\epsilon_0$, when finding the first eigenvector, so the overall accuracy of all eigenvectors maintains at the claimed level.
}

\begin{remark}\label{rem:complex}
{In a rare scenario, $\lambda$ may exactly be at the middle of two eigenvalues, so  $\arg\min_i|\lambda-\lambda_i|$ is not a single point. In this case, we can re-run the algorithm with a slightly larger and smaller $\lambda$.
}
\end{remark}

\subsection{FLICKER: Federated Learning of Iterative Communal Kernels for Eigenvector Recovery}
We now extend Algorithm~\ref{alg:GD} from a symmetric matrix to a rectangular data matrix $B\in\mathbb R^{n\times m}$. Let
$B=U\Sigma V^\top$ be its singular value decomposition (SVD). The left and right singular vectors of $B$ are the eigenvectors of $BB^\top$ and $B^\top B$, respectively, with corresponding eigenvalues $\sigma_i(B)^2$.
{Algorithm \ref{alg:GD} can be applied to $B$ with input target values as $\lambda = \sigma^2$ and iterations on $BB^\top v^t$ or $B^\top B v^t$ to recover the singular vectors on both sides.}

In modern applications, the full matrix $B$ is not directly available. Instead, there are $K$ local servers and each has a local data matrix $B_k \in \reals^{n_k \times m}$, where $n = \sum_{k=1}^K n_k$. The full data matrix is a stack of these matrices that $B = [B_1^\top, B_2^\top, \cdots, B_K^\top]^\top \in \reals^{n \times m}$.
Due to communication costs and privacy concerns, these local data servers communicate only the processed results, rather than the original data.
{Naturally, our goal is to find the top $L$ left and right singular vectors of $B$ based on $B_k$'s, without directly stacking them together.}

Denote $\sigma_1(B) \geq \sigma_2(B) \geq \cdots \geq \sigma_L(B)$ be {the top $L$} singular values of $B$ and $u_i$ be the corresponding left singular vector. Hence, $u_i$ is the eigenvector of $BB^\top$ corresponding to the eigenvalue $\lambda_i = \sigma_i(B)^2$. Apply Algorithm \ref{alg:GD} with $A = BB^\top$ and a tuning parameter $\sigma^2$ close to $\sigma_i(B)^2$ to recover $u_i$ if we have the full data $B$.

When only $B_k$ are available, we want to revise Algorithm \ref{alg:GD} to adapt to this federated setting. The main idea is to calculate the update
\((\sigma^2 I - BB^\top)^2 v^t = \sigma^4 v^t - 2\sigma^2 BB^\top v^t + BB^\top BB^\top v^t\) based on $B_k$.
We partition $v^t$ into $v^t=[(v^t_{(1)})^\top, \ldots,
(v^t_{(K)})^\top]^\top$, where $v^t_{(k)}\in\mathbb{R}^{n_k}$ corresponds to the sample indices on local server $k$.
Therefore, $BB^\top v^t$ follows that:
\begin{equation}\label{eqn:bkv}
BB^\top v^t =
\begin{bmatrix}
B_1\\
\vdots\\
B_K
\end{bmatrix}
[B_1^\top, B_2^\top, \cdots, B_K^{\top}]
\begin{bmatrix}
v^t_{(1)}\\
\vdots\\
v^t_{(K)}
\end{bmatrix}
=
\begin{bmatrix}
B_1 x_t\\
\vdots\\
B_K x_t
\end{bmatrix},
\quad \mbox{where }
x_t := \sum_{k=1}^K B_k^{\top}v^t_{(k)}.
\end{equation}
Hence, $BB^\top v^t$ can be computed using only the local
matrix--vector products $B_k^\top v^t_{(k)}$ and $B_kx^t$; applying the same scheme once more yields $(BB^\top)^2v^t$.
Therefore, we can update $v^{t+1}$ using only matrix-vector products from local servers. It largely reduces the privacy concern and communication cost.

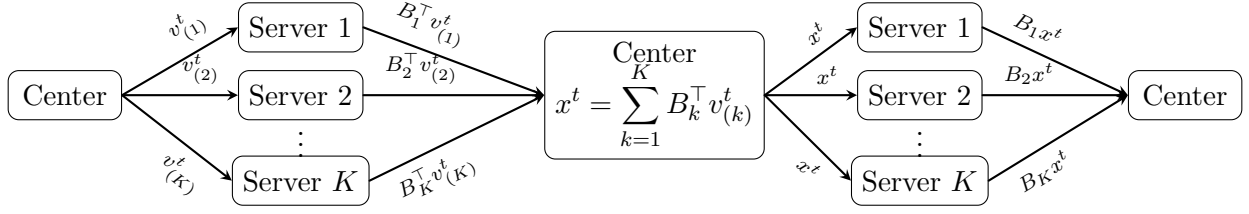
\begin{figure}
\centering
\begin{tikzpicture}[
    >=stealth,
    line cap=round,
    scale=0.78,
    every node/.style={font=\small},
    box/.style={
        draw,
        rounded corners,
        align=center,
        minimum width=1.5cm,
        minimum height=0.65cm
    },
    flow/.style={
        ->,
        thick
    }
]

\def\yOne{1.4}
\def\yTwo{0.25}
\def\yDots{-0.45}
\def\yK{-1.2}

\def\xcA{0}
\def\xsA{4.0}
\def\xcB{10.0}
\def\xsB{14.5}
\def\xcC{19.0}

\node[box] (C1) at (\xcA,\yTwo) {Center};

\node[box, minimum width=2.1cm] (C2) at (\xcB,\yTwo)
{Center\\[-1pt]
$\displaystyle x^t=\sum_{k=1}^K B_k^\top v^t_{(k)}$};

\node[box] (C3) at (\xcC,\yTwo) {Center};

\node[box] (S1A) at (\xsA,\yOne) {Server 1};
\node[box] (S2A) at (\xsA,\yTwo) {Server 2};
\node       (DotsA) at (\xsA,\yDots) {$\vdots$};
\node[box] (SKA) at (\xsA,\yK) {Server $K$};

\node[box] (S1B) at (\xsB,\yOne) {Server 1};
\node[box] (S2B) at (\xsB,\yTwo) {Server 2};
\node       (DotsB) at (\xsB,\yDots) {$\vdots$};
\node[box] (SKB) at (\xsB,\yK) {Server $K$};

\draw[flow] (C1.east) -- (S1A.west)
node[pos=0.68, above, sloped] {\scriptsize $v^t_{(1)}$};

\draw[flow] (C1.east) -- (S2A.west)
node[pos=0.68, above] {\scriptsize $v^t_{(2)}$};

\draw[flow] (C1.east) -- (SKA.west)
node[pos=0.68, below, sloped] {\scriptsize $v^t_{(K)}$};

\draw[flow] (S1A.east) -- (C2.west)
node[pos=0.32, above, sloped] {\scriptsize $B_1^\top v^t_{(1)}$};

\draw[flow] (S2A.east) -- (C2.west)
node[pos=0.32, above] {\scriptsize $B_2^\top v^t_{(2)}$};

\draw[flow] (SKA.east) -- (C2.west)
node[pos=0.32, below, sloped] {\scriptsize $B_K^\top v^t_{(K)}$};

\draw[flow] (C2.east) -- (S1B.west)
node[pos=0.68, above, sloped] {\scriptsize $x^t$};

\draw[flow] (C2.east) -- (S2B.west)
node[pos=0.68, above] {\scriptsize $x^t$};

\draw[flow] (C2.east) -- (SKB.west)
node[pos=0.68, below, sloped] {\scriptsize $x^t$};

\draw[flow] (S1B.east) -- (C3.west)
node[pos=0.32, above, sloped] {\scriptsize $B_1x^t$};

\draw[flow] (S2B.east) -- (C3.west)
node[pos=0.32, above] {\scriptsize $B_2x^t$};

\draw[flow] (SKB.east) -- (C3.west)
node[pos=0.32, below, sloped] {\scriptsize $B_Kx^t$};

\end{tikzpicture}

    \caption{Illustration of recovering $BB^\top v^t$ from $v^t$ in the federated setting.}
    \label{fig:alg1}
\end{figure}

Based on \eqref{eqn:bkv}, we propose FLICKER for the federated eigenvector recovery in Algorithm \ref{alg:decent1}. The steps for one iteration can be found in Figure \ref{fig:alg1}, where the center server broadcasts and stacks vectors and the local servers return matrix-vector products. {Furthermore, with a left singular vector $u$, its singular value is given by $\sqrt{u^\top BB^\top u}$. Hence, we also return an estimate of the singular value with $\sqrt{u^\top BB^\top u}$.}

{We next propose Algorithm \ref{alg:decent2} for the right singular vectors $v_i$. Here we use $v_i$ to denote right singular vectors and $u^t$ for $t$-th iteration inside an algorithm. This scenario is actually simpler. To find the right singular vector, we should apply Algorithm \ref{alg:GD} to $B^\top B$ with a target value $\sigma^2$.
Hence, the update in each iteration is about $B^\top B u^t$, where
\begin{equation}\label{eqn:bku}
B^\top B u^t =\sum_{k=1}^K B_k^{\top} B_k u^t
\end{equation}
Therefore, the required quantity $B^\top B u^t$ can be found by aggregating $B_k^{\top} B_k u^t$ from each local server. It follows for $(B^{\top} B)^2 u^t$ and hence Algorithm \ref{alg:decent2}.
}

{Finally, we investigate FLICKER} in terms of its {communication and} computation cost. We take Algorithm \ref{alg:decent1}, the more complicated algorithm, for discussion. Each gradient descent iteration requires two federated products, $BB^\top v^t$ and $(BB^\top)^2 v^t$.
The communication costs and computational costs for each federated product are summarized below.
\begin{itemize}
  \item \emph{Reduce.} Each server \(k\) sends a vector in \(B_k^\top v^t_{(k)} \in \reals^m\) to the center, for a total of \(Km\) scalars. The computation cost on each local server is $O(n_km)$.
  \item \emph{Broadcast.} The center sends the aggregated \(\mathbb{R}^m\) vector back to all \(K\) servers, for another \(Km\) scalars. The computation cost is $O(Km)$.
  \item \emph{Boundary vectors.} At the start and end of each iteration, the center broadcasts \(v^t\in\reals^n\) and receives the concatenated vectors
  \(\{u^t_{(k)}\}\), contributing \(O(n)\) scalars per boundary exchange.
\end{itemize}
Thus the total communicated payload per iteration is \(O(3n + 4Km)\) scalars.
On server \(k\), each local product \(B_k x\) or \(B_k^{\top} y\) costs \(O(n_k m)\) operations in the dense case, and \(O(|E_k|)\) operations when \(B_k\) is a sparse matrix with \(|E_k|\) nonzero entries.
By Theorem~\ref{thm:GD}, the number of iterations needed is \(T = O(\log n)\) \blue{when $r$ is uniformly bounded away from 1 for fixed accuracy and failure probability}, so the total communication scales as
\(O\big((3n + 4Km)\log n\big)\) and the total local computation scales as
\(O(|E_k|\log n)\).
{To recover top $L$ singular vectors, the same procedure needs to be repeated $L$ times using either singular value estimates or the grid search procedure in Algorithm \ref{alg:lambda} below. This will bring the total local computation scales as \(O(L|E_k|\log n)\) for the top $L$ singular vectors or \(O\big(\min\{n,m\}|E_k|)\log n\big)\) for the full spectrum.
}

As a comparison, if we gather all data to the center and perform SVD, then the total communication cost is of $O(|E|)$, where $|E|=\sum_{k=1}^K |E_k|$.
The computation cost with the same iterative spectral procedure centrally requires $O(L|E|\log n)$ for top $L$ singular directions.
Algorithm~\ref{alg:decent1} retains the same order of total computational cost, while reducing the communication cost to
$O((n+Km)\log n)$, which can be substantially smaller when $n+Km\ll |E|$.
Moreover, the dominant computations are carried out at the local servers, whereas the center only performs vector aggregation and updates. This architecture is particularly attractive when the local data holders have substantially greater computational resources than the coordinating center.

Federated learning provides better privacy protection than centralized learning. In FLICKER, the local servers only need to release $B_k^\top x$ or $B_k x$ for a given vector $x$. So an attacker may learn information of $B_k$ along some directions but not the whole matrix. \blue{In spectral community detection}, we usually require only the leading $L \ll \min\{n,m\}$ singular vectors instead of the full spectrum, so that the privacy leakage is limited.
However, this is not a formal privacy guarantee.
Formal differential privacy requires explicit randomization and a sensitivity-based privacy--accuracy analysis \citep{wang2026privacy}. Such guarantees have
been developed for spectral community detection in \cite{wang2026netptr}; incorporating them into the federated setting is beyond the scope of this paper.

The same design principle can also adapt other eigenvector  algorithms to the federated setting. For example, inverse iteration \citep{golub2013matrix} and Krylov subspace methods
\citep{saad2011numerical} rely on repeated matrix--vector products, $\operatorname{span}\{v^t,Av^t,\ldots,A^Jv^t\}$. These methods can be implemented in the federated setting using the communication schemes in \eqref{eqn:bkv} and \eqref{eqn:bku}. By contrast, our method requires only $Ax$ and $A^2x$ for a given vector $x$ at each iteration, leading to a more straightforward federated implementation.

\begin{algorithm}[!ht]
\small
\caption{Federated power iteration FLICKER (left singular vectors)}
\label{alg:decent1}
\begin{algorithmic}
\Require Local blocks $B_k$ on servers $k=1,\ldots,K$; approximate singular value $\sigma$; number of iterations $T$; step size $\eta$.
\Ensure {$(\sigma_v,v^T)$: left singular value and vector of $B$ where $\sigma_v$ is closest to $\sigma$.}
\State Initialize $v^0 \sim \mathcal{N}(0,I_n)$ and set $v^0 \leftarrow v^0 / \|v^0\|$.
\For{$t = 0,\ldots,T-1$}
  \State [Center] Send the block $v^t_{(k)}$ to server $k$.
  \For{$k = 1,\ldots,K$}
    \State [Server $k$] Compute $x_{(k)}^t = B_k^\top v^t_{(k)}$ and send $x_{(k)}^t$ to the center.
  \EndFor
  \State [Center] Aggregate $x^t = \sum_{k=1}^K x_{(k)}^t$ and broadcast $x^t$.
  \For{$k = 1,\ldots,K$}
    \State [Server $k$] Compute $u^t_{(k)} = B_k x^t$ and send $u^t_{(k)}$ to the center.
  \EndFor
  \State [Center] Concatenate $u^t = [(u^t_{(1)})^\top,\ldots,(u^t_{(K)})^\top]^\top$.
   \State {[Center] Update $\sigma_v = \sqrt{\max\{( u^t)^\top v^t,0\}}$.}
  \For{$k = 1,\ldots,K$}
    \State [Server $k$] Compute $y_k^t = B_k^\top u^t_{(k)}$ and send $y_k^t$ to the center.
  \EndFor
  \State [Center] Aggregate $y^t = \sum_{k=1}^K y_k^t$ and broadcast $y^t$.
  \For{$k = 1,\ldots,K$}
    \State [Server $k$] Compute $w^t_{(k)} = B_k y^t$ and send $w^t_{(k)}$ to the center.
  \EndFor
  \State [Center] Concatenate $w^t = [(w^t_{(1)})^\top,\ldots,(w^t_{(K)})^\top]^\top$.
  \State [Center] Update
        $v^{t+1} = v^t - \eta (\sigma^4 v^t - 2\sigma^2 u^t + w^t)$, and set
        $v^{t+1} \leftarrow \frac{v^{t+1}}{\|v^{t+1}\|}$.

\EndFor
\end{algorithmic}
\end{algorithm}

\begin{algorithm}[!t]
\small
\caption{Federated power iteration FLICKER (right singular vectors)}
\label{alg:decent2}
\begin{algorithmic}
\Require $B_k$, $k=1,\ldots,K$, target singular value $\sigma$, number of iterations $T$, step size $\eta$.
\Ensure {$(\sigma_v, u^T)$: right singular value and vector of $B$ with $\sigma_v$ closest to $\sigma$. }
\State Initialize $u^0 \sim \mathcal{N}(0,I_m)$ and set $v^0 \leftarrow u^0 / \|u^0\|$.
\For{$t = 0,\ldots, T-1$}
  \State [Center] Broadcast $u^{t}$.
  \For{$k = 1,\ldots, K$}
    \State [Server $k$] Compute $x^{t}_{k}= B_k^\top B_k u^t$ and send $x^t_k$ to center.
  \EndFor
  \State [Center] Sum $x^{t} = \sum_{k=1}^K x^t_k$ and broadcast $x^t$.
   \State {[Center] Update $\sigma_v = \sqrt{\max\{( x^t)^\top u^t,0\}}$.}
  \For{$k = 1,\ldots, K$}
    \State [Server $k$] Compute $w^{t}_{k}= B_k^\top B_k x^t$ and send $w^t_k$ to center.
  \EndFor
  \State [Center] Sum $w^{t} = \sum_{k = 1}^K w^t_k$.
  \State [Center] Update
    $
      u^{t+1} = u^t - \eta \big(\sigma^4 u^{t} - 2\sigma^2 x^{t} + w^{t}\big),
      \qquad
      u^{t+1} \leftarrow {u^{t+1}}/{\|u^{t+1}\|}.
    $
\EndFor
\end{algorithmic}
\end{algorithm}

\subsection{Selection on target singular values}\label{sec:lambda}
Given a target value $\sigma$, FLICKER recovers a singular direction whose squared singular value is closest to $\sigma^2$. When proper approximations of the target singular values are available, different singular directions can be recovered independently, without sequential orthogonalizations {used in standard power methods}. This raises the practical question of how to select the target values.

When an approximate target is available, it can be used directly in FLICKER. For example, the leading singular direction can be recovered using any target above the largest singular value. {For a binary bipartite adjacency matrix, $\|B\|_F=\sqrt{|E|}$ provides such an upper bound.} Another possibility is to use local spectral information. In
applications where the leading singular values scale with the local sample size, a properly rescaled local singular value can provide an effective target for FLICKER. This is the strategy used in Sections \ref{sec:simu}--\ref{sec:data}.

When we would like to recover the full spectrum, we can instead
search over the spectrum through repeated calls to FLICKER.  Suppose that upper and lower tuning values $\sigma^{(1)}$ and $\sigma^{(2)}$ bracket the singular values of interest. Running \gdpi at the two endpoints first identifies the largest and smallest singular directions. We then recursively examine the intervals between the singular values already discovered.
{For a general bipartite network setting, we may  choose $\sigma^{(1)}=\sqrt{\text{total degree}}$ and $\sigma^{(2)}=0$.}

To search for additional singular vectors, we recursively bisect intervals between already discovered approximate singular values. Suppose we already recover two consecutive singular-value estimates $\hat{\sigma}^{(i)}>\hat{\sigma}^{(i+1)}$, we use the target value
$\sigma=\sqrt{\{(\hat{\sigma}^{(i)})^2+(\hat{\sigma}^{(i+1)})^2\}/2}.$
We then compute $(\hat{\sigma},u)=\mathrm{FLICKER}(\sigma)$. If $u$ is approximately orthogonal to both $v^{(i)}$ and $v^{(i+1)}$, it represents a new singular direction; we insert $(\hat{\sigma},u)$ between the two existing pairs and continue the search on the two resulting intervals.
Otherwise, the current interval is marked as completed. The resulting procedure is summarized in Algorithm~\ref{alg:lambda}.
Although the procedure can recover the full spectrum, in practice it can be terminated once the desired number of singular directions has been obtained.

\begin{algorithm}[!t]
\small
\caption{Grid search for singular values and singular vectors}\label{alg:lambda}
\begin{algorithmic}
\Require \gdpi algorithm, upper and lower spectral bounds
$\sigma^{(1)}$, $\sigma^{(2)}$.
\Ensure A sequence of approximate singular pairs
{$\{(\hat{\sigma}^{(i)}, v^{(i)})\}$}.

\State {Compute two initial approximate singular pairs}
       ${(\hat{\sigma}^{(1)},v^{(1)})} = \gdpi(\sigma^{(1)})$
       and
       ${(\hat{\sigma}^{(2)},v^{(2)})} = \gdpi(\sigma^{(2)})$.

\State Initialize singular value list
       ${I = [\hat{\sigma}^{(1)}, \hat{\sigma}^{(2)}]}$
       (sorted in decreasing order).
\State Initialize singular vector list $V = [v^{(1)}, v^{(2)}]$.

\State {Initialize target value list
       $J =
       \left[
       \sqrt{\tfrac12\{(\hat{\sigma}^{(1)})^2+
                         (\hat{\sigma}^{(2)})^2\}}
       \right]$.}

\While{$J$ is not empty and $|I|\leq \min\{m,n\}$}
  \State Select a $\sigma \in J$ and remove it from $J$.
  \State Find $i$ such that
         {$\hat{\sigma}^{(i)} > \sigma >
         \hat{\sigma}^{(i+1)}$}, where
         $\hat{\sigma}^{(i)}$ and $\hat{\sigma}^{(i+1)}$
         are consecutive elements in $I$.

  \State Compute {$(\hat{\sigma},u) = \gdpi(\sigma)$}.

  \If{$|u^\top v^{(i)}| \leq 1/2$ \textbf{ and }
      $|u^\top v^{(i+1)}| \leq 1/2$}

    \State Insert {$\hat{\sigma}$} into $I$ between
           $\hat{\sigma}^{(i)}$ and $\hat{\sigma}^{(i+1)}$.
    \State Insert $u$ into $V$ between $v^{(i)}$ and $v^{(i+1)}$.

    \State {Append the new target values
           $\sqrt{\tfrac12\{(\hat{\sigma}^{(i+1)})^2+\hat{\sigma}^2\}}$
           and
           $\sqrt{\tfrac12\{(\hat{\sigma}^{(i)})^2+\hat{\sigma}^2\}}$
           to $J$.}
  \EndIf
\EndWhile
\end{algorithmic}
\end{algorithm}

{
The next proposition establishes the grid-search guarantee of the full spectrum when singular values are simple. When a discovered singular value has multiplicity greater than one, the grid search in Algorithm \ref{alg:lambda} identifies its spectral location, and then the projected version of FLICKER in Algorithm~\ref{alg:GD} can be applied sequentially with the same target value to recover the corresponding singular subspace.
\begin{proposition}
Suppose that $B$ has $\min\{m,n\}$ nonzero and simple singular values, and that the initial target values $\sigma^{(1)}>\sigma_1\geq \sigma_{\min\{m,n\}}>\sigma^{(2)}$.
Assume further that, whenever an interval examined by
Algorithm~\ref{alg:lambda} contains an undiscovered singular value, the squared-midpoint target value has a unique closest singular value in the interior of the interval. If each FLICKER call is run to convergence, then Algorithm~\ref{alg:lambda} recovers all singular value--vector pairs, with at most $2\min\{m,n\}-1$ calls to FLICKER.
\end{proposition}
}

\section{Federated Community Detection}\label{sec:fedcd}
We now formalize the federated bipartite network setting.
Let $\mathcal L$ and $\mathcal R$ denote the left and right node sets, with $n=|\mathcal L|$ and $m=|\mathcal R|$, and let $B\in\{0,1\}^{n\times m}$ be the whole bipartite adjacency matrix.
In the federated setting, the left node set is partitioned as $\mathcal L=\bigcup_{k=1}^K\mathcal L^{(k)}$, where the $\mathcal L^{(k)}$'s are disjoint.
Server $k$ observes the interactions between $\mathcal L^{(k)}$ and the common node set $\mathcal R$, and hence stores $B_k\in\{0,1\}^{n_k\times m}$, where $n_k=|\mathcal L^{(k)}|$, $\sum_{k=1}^K n_k=n$, and $B=[B_1^\top,\ldots,B_K^\top]^\top$.
Let $z_1\in[L_1]^n$ and $z_2\in[L_2]^m$ denote the community labels
on the two node sets. Our goal is to recover $z_1$ and $z_2$ from
$\{B_k\}_{k=1}^K$ without pooling the local matrices.

We adopt the bipartite stochastic block model (SBM)
\citep{specbi,disim}. Let $Z_1\in\{0,1\}^{n\times L_1}$ and $Z_2\in\{0,1\}^{m\times L_2}$ be the membership matrices corresponding to $z_1$ and $z_2$, and let $G\in[0,1]^{L_1\times L_2}$ be the community connection matrix. Then
\[
Q:=\mathbb E[B]=Z_1GZ_2^\top.
\]
Let $Z_1^{(k)}$ denote the rows of $Z_1$ indexed by
$\mathcal L^{(k)}$. The corresponding local mean matrix is
$Q_k:=\mathbb E[B_k]=Z_1^{(k)}GZ_2^\top$, $1\le k\le K$.

To recover $z_1$ and $z_2$, \cite{specbi} applies SVD to $B$ and clusters the two node sets using its leading singular vectors.
Related spectral methods include \cite{disim}, which uses regularization to accommodate degree heterogeneity, and \cite{Dscore}, which uses entrywise ratios between leading singular vectors and the first singular vector.
These methods rely on the leading singular vectors of $B$, and thus cannot be directly implemented in our federated setting, where only the local matrices $B_k$ are observed.

We next introduce BiFLICKER, which combines our \gdpi algorithm with a simple singular-value selection scheme. To recover the leading singular vectors of $B$, we first estimate the corresponding singular values from one local server. Proposition~\ref{prop:local-gram-calibration} below shows that, after rescaling, the leading singular values of a local block $B_k$ are close to those of the global matrix $B$. Moreover, FLICKER only requires the input target value $\sigma$ to be closer to the singular value $\sigma_i(B)$ of our interest, than any other singular values, on the squared singular-value scale. It does not require the exact value. We therefore compute the leading local singular values $\sigma_1^{(k)},\ldots,\sigma_L^{(k)}$, where $L=\min\{L_1,L_2\}$, on any server $k$ with sufficiently large $n_k$, and use $\tilde{\sigma}_i=\sqrt{n/n_k}\,\sigma_i^{(k)}$, $i=1,\ldots,L$, as target values for FLICKER.
In this $\sigma_i$ estimation step, a sufficiently large reference server is preferable.
The resulting left and right singular
vectors of $B$ are then used as embeddings for $k$-means clustering.
The procedure is summarized in Algorithm~\ref{alg:bi}.

\begin{algorithm}[htbp]
\caption{Federated community detection on bipartite networks (BiFLICKER)}
\label{alg:bi}
\begin{algorithmic}
\Require Local connection matrices $B_k$, {total number of left nodes $n$},
        numbers of communities $L_1$ and $L_2$ for
        $\mathcal{L}$ and $\mathcal{R}$, iterations $T$, step size $\eta$.
\Ensure {Estimates $\hat z_1$ and $\hat z_2$ of the community labels
        $z_1$ and $z_2$.}
\State [{Any server $k$ with sufficiently large $n_k$}] Compute its leading singular values
       $\sigma_1^{(k)} \geq \sigma_2^{(k)} \geq \cdots \geq \sigma_L^{(k)}$, where
       $L = \min\{L_1,L_2\}$, and send $\{n_k,\sigma_i^{(k)}\}_{i=1}^L$ to the center.
\State [Center] Form the {target values}
       $\mathcal{I} = ({\tilde{\sigma}_1},\dots,{\tilde{\sigma}_L})$ with
       ${\tilde{\sigma}_i} = \sqrt{n/n_k}\,\sigma_i^{(k)}$.

 \State {For each target $\tilde{\sigma}_i\in\mathcal I$, independently apply the federated left- and right-singular-vector FLICKER algorithms to obtain $\xi_i$ and $\psi_i$, respectively. Set $\Xi=[\xi_1,\ldots,\xi_L]$ and $\Psi=[\psi_1,\ldots,\psi_L]$.} See remark on a variant Sequential BiFLICKER.
\State Apply $k$-means to the rows of $\Xi$ to cluster the $n$ nodes in $\mathcal{L}$
       into $L_1$ groups; the resulting labels form $\hat{z}_1$.
\State Apply $k$-means to the rows of $\Psi$ to cluster the $m$ nodes in $\mathcal{R}$
       into $L_2$ groups; the resulting labels form $\hat{z}_2$.
\end{algorithmic}
\end{algorithm}
\begin{remark}
When the target singular values are close or repeated, we consider a sequential variant, {\it Sequential BiFLICKER}.
Starting from $\tilde{\sigma}_1$, after recovering $\xi_1,\ldots,\xi_{i-1}$ and $\psi_1,\ldots,\psi_{i-1}$, define $P_i^{\mathcal L}=I-\Xi_{<i}\Xi_{<i}^\top$ and
$P_i^{\mathcal R}=I-\Psi_{<i}\Psi_{<i}^\top$.
Then apply Algorithm~\ref{alg:decent1} with target $\tilde\sigma_i$ and projection $P_i^{\mathcal L}$ to obtain $\xi_i$, and Algorithm~\ref{alg:decent2} with target $\tilde\sigma_i$ and projection $P_i^{\mathcal R}$ to obtain $\psi_i$, where we apply the projected FLICKER.
This sequential projection is more robust to nearby singular values. It is used in Theorem \ref{thm:sequential-community-recovery} and real data analysis in Section~\ref{sec:data}.
\end{remark}
\begin{remark}
    \blue{$L$ and the number of communities $L_1$ and $L_2$ are input parameters. When they are unknown, one can first choose a working embedding dimension $L$ using spectral diagnostics, then select $L_1$ and $L_2$ separately on the fixed estimated embeddings using $k$-means diagnostic statistics. A practical heuristic is to maximize the mean Euclidean silhouette over $k$-means partitions, which is applied in the Supplementary Material.}
\end{remark}

\section{Theoretical Guarantee}\label{sec:theory}
This section presents the theoretical guarantees for FLICKER and BiFLICKER. Section~\ref{subsec:flicker}  provides the proof of Theorem~\ref{thm:GDproj}; the proof of Theorem~\ref{thm:GD} is simpler and deferred to the supplementary material. Section~\ref{subsec:strong-consistency} establishes the community recovery guarantees for BiFLICKER under the bipartite SBM.

\subsection{Convergence of FLICKER}
\label{subsec:flicker}
Here we demonstrate the proof for our FLICKER algorithm with a projection $P$. Starting with the exact projection $P^\perp_I$, a spectral decomposition shows that the iteration approaches the target subspace at a geometric rate. We then bound the discrepancy between the exact projection and the given projection $P$. A lower bound on the initial target component under Gaussian initialization controls the effect of normalization, allowing us to combine the convergence and perturbation estimates to obtain the claimed bound.

{
\begin{proof}[Proof of Theorem~\ref{thm:GDproj}]
Let $\mu_\star=\max_{j\notin I}\{1-\eta(\lambda-\lambda_j)^2\}$ and
set $M=\mu_\star^{-1}[I_n-\eta(\lambda I_n-A)^2]$.
The step-size condition implies $M\succ0$ and $\|M\|\le\mu_\star^{-1}$.
Since $P$ is a nonzero orthogonal projector, $PMv^0\ne0$ almost surely.
Moreover, for any nonzero $x\in\operatorname{range}(P)$,
$x^\top PMx=x^\top Mx>0$, so subsequent updates are also nonzero.
Thus, setting $w^t=(PM)^t v^0$ and $z^t=(P_I^\perp M)^t v^0$,
we have $v^t=w^t/\|w^t\|$ almost surely for every $t\ge1$.

We first consider the exact projection $P_I^\perp$.
Recall that $P_I^\perp=I_n-\sum_{i\in I}u_i u_i^\top$ is the exact orthogonal projector onto the remaining eigendirections.
The matrix $P_I^\perp M$ is zero on the excluded directions, is the identity
on $\mathcal U_\star$, and has eigenvalues at most $r_I$ on the other
remaining directions. Consequently, $\|P_I^\perp M\|=1$ and, for every
$t\ge1$, we have $\|z^t\|\ge\|P_\star v^0\|$ and
$\|(I_n-P_\star)z^t\|\le r_I^t$.
For a uniform unit vector $v^0$ and any fixed unit vector $u$,
$\Pr\{|u^\top v^0|<\varsigma/\sqrt n\}\le\varsigma$.
Taking $u\in\mathcal U_\star$, we therefore have
$\|P_\star v^0\|\ge\varsigma/\sqrt n$
with probability at least $1-\varsigma$.
We work on this event below.
In particular,
\[
  \operatorname{dist}\!\left(\frac{z^T}{\|z^T\|},\mathcal U_\star\right)
  \le\frac{\sqrt n}{\varsigma}r_I^T.
\]

Next, we focus on the given projection $P$. Note that $\|(P-P_I^\perp)M\|\le\epsilon_0/\mu_\star$ and hence
$\|PM\|\le1+\epsilon_0/\mu_\star$.
Telescoping the difference of the two matrix powers gives
\[
  \|w^T-z^T\|
  \le\frac{T\epsilon_0}{\mu_\star}
       \left(1+\frac{\epsilon_0}{\mu_\star}\right)^{T-1}.
\]
If $T\epsilon_0/\mu_\star\le1$, the right-hand side is at most
$eT\epsilon_0/\mu_\star$.
For nonzero $x,y$, normalization satisfies
$\bigl\|x/\|x\|-y/\|y\|\bigr\|\le2\|x-y\|/\|y\|$.
Applying this with $x=w^T$ and $y=z^T$, and using the triangle inequality,
we obtain
\[
  \operatorname{dist}(v^T,\mathcal U_\star)
  \le\frac{\sqrt n}{\varsigma}r_I^T
     +\frac{2e\sqrt n\,T\epsilon_0}{\varsigma\mu_\star}.
\]
If $T\epsilon_0/\mu_\star>1$, the same inequality holds because its
right-hand side exceeds one, whereas
$\operatorname{dist}(v^T,\mathcal U_\star)\le1$.

For $0<r_I<1$, the first term is at most $\epsilon_1$ provided
$r_I^T\le\varsigma\epsilon_1/\sqrt n$.
This is ensured by the stated iteration requirement after increasing
$C_\varsigma$ if necessary. When $r_I=0$, the first term vanishes for
every $T\ge1$. Taking $C_\varsigma\ge2e/\varsigma$ proves the result.
\end{proof}
}

\subsection{Consistency of BiFLICKER}\label{subsec:strong-consistency}
We now show that the theoretical guarantee of Sequential BiFLICKER under the bipartite SBM introduced in Section~\ref{sec:fedcd}.
Let $z_1:[n]\to[L_1]$ and $z_2:[m]\to[L_2]$ be fixed community labels.
Under the bipartite stochastic block model, the entries of $B$ are independent, with $B(i,j)\sim\operatorname{Bernoulli}(G(z_1(i),z_2(j)))$. Write $Q=\mathbb E[B]=Z_1GZ_2^\top$ and $L=\min\{L_1,L_2\}$. Recall that $Q_k = \mathbb{E}[B_k]$ on server $k$.
We impose the following assumption on the model.
{
\begin{assumption}[Community proportions and connection probabilities]
\label{aspt:Bipart}
For a constant $\rho\in(0,1)$ and a density parameter
$\theta_0 = \theta_0(n,m)\in(0,1]$, which may depend on $n$ and $m$, suppose that:
\begin{enumerate}
\item[(i)] The proportion of each community on either side is at least $\rho$, uniformly in $n$ and $m$.
\item[(ii)] The connection probabilities satisfy $0\leq G(l_1,l_2)\leq\theta_0(n,m)$ for all $l_1\in[L_1]$ and $l_2\in[L_2]$, and $\sigma_L(G)\geq\rho\theta_0(n,m)$.
\item[(iii)] For any distinct $l_1,l_1'\in[L_1]$,
$\max_{l_2\in[L_2]}|G(l_1,l_2)-G(l_1',l_2)|\geq\rho\theta_0(n,m)$.
Similarly, for any distinct $l_2,l_2'\in[L_2]$,
$\max_{l_1\in[L_1]}|G(l_1,l_2)-G(l_1,l_2')|\geq\rho\theta_0(n,m)$.
\end{enumerate}
\end{assumption}
}
{Condition~(i) prevents vanishing community proportions.
Condition~(ii) specifies the density scale and ensures a nondegenerate rank-$L$ signal, while condition (iii) separates the connection profiles of distinct communities on both sides. $\theta_0$ here should be interpreted as the density level parameter. It may decay with an increasing $n$, representing the sparse network cases.
Singular values with multiplicity larger than 1 are allowed.
}

\begin{proposition}[Local spectral calibration]
\label{prop:local-gram-calibration}
Suppose Assumption~\ref{aspt:Bipart} holds.
Suppose there is a local server $k$ with $n_k\geq\rho n$, uniformly sampled from $n$ nodes.
For a fixed $\alpha>1$, suppose $m\leq n^\alpha$ and  {$\min\{n,m\}\theta_0/\log n$ is} sufficiently large.
Then there is a constant $C>0$ such that, with probability tending to one,
\begin{equation}
\label{eq:local-gram-calibration}
\left|(n/n_k)\sigma_j(B_k)^2-\sigma_j(B)^2\right|
\leq C\left\{
\theta_0^2m\sqrt n\,\log n
+\theta_0^{3/2}\sqrt{nm(n+m)}
\right\}, \quad 1 \leq j \leq L.
\end{equation}
\end{proposition}

Uniform sampling is one sufficient condition for local calibration.
The recovery result below applies to any reference block satisfying
\eqref{eq:local-gram-calibration} with probability tending to one;
it imposes no representative-composition requirement on the other servers.

\begin{theorem}[Community recovery by sequential BiFLICKER]
\label{thm:sequential-community-recovery}
Suppose Assumption~\ref{aspt:Bipart} holds, $m\leq n^\alpha$ for a fixed $\alpha>1$, and {$\min\{n,m\}\theta_0/\log n$ is} sufficiently large.
Let $B_k$ be a reference block satisfying \eqref{eq:local-gram-calibration} with probability tending to one.

Apply Sequential BiFLICKER, the variant of Algorithm \ref{alg:bi} in Remark~2,
with targets $\widetilde\sigma_j=\sqrt{n/n_k}\,\sigma_j(B_k)$, $j=1,\ldots,L$, step sizes $\eta_j\asymp(nm\theta_0^2)^{-2}$  such that $\eta_j\max\{\widetilde\sigma_j^4,
(\sigma_1(B)^2-\widetilde\sigma_j^2)^2\}\leq1-c_1$
for a fixed $c_1\in(0,1)$, and $T=\lceil C_T\log(n+m)\rceil$ iterations per component, for a
sufficiently large fixed $C_T$.
Then, with high probability, $\widehat{z}_1$ and $\widehat{z}_2$ satisfy that
\begin{equation}
\label{eq:sequential-misclassification}
\max\left\{
\min_{\pi_1\in\mathfrak S_{L_1}}\frac1n
\sum_{i=1}^n\mathbf1\{\widehat z_1(i)\ne\pi_1(z_1(i))\},\;
\min_{\pi_2\in\mathfrak S_{L_2}}\frac1m
\sum_{j=1}^m\mathbf1\{\widehat z_2(j)\ne\pi_2(z_2(j))\}
\right\}
\leq C\frac{n+m}{nm\theta_0},
\end{equation}
where $\mathfrak S_r$ denotes the permutations of $[r]$.
If, in addition, {$\min\{m,n\}\theta_0^2/\log n$} is sufficiently large, both labelings are recovered exactly up to separate permutations with high probability.
\end{theorem}

Theorem~\ref{thm:sequential-community-recovery} states that, under Assumption~\ref{aspt:Bipart}, the BiFLICKER embedding followed by $k$-means achieves
{$o(1)$ misclassification rate in sparse networks and exact community recovery under a stronger sparsity condition.} In particular, {once the rescaled local targets identify the correct singular directions, BiFLICKER guarantees weak consistency whenever $\min\{n,m\}\theta_0\to\infty$, matching the
weak-consistency sparsity scale in \cite{zhou2019analysis}. The errors introduced by federated iterations are asymptotically negligible.}

{
Our strong-consistency analysis leverages $2\to\infty$ eigenvector error bounds \citep{denoising} and uses techniques related to \cite{su2019strong} and \cite{abbe2020entrywise}. Compared with the centralized symmetric settings considered in these works, our analysis requires several additional steps. We need to control both the left and right singular spaces of the rectangular bipartite matrix, justify the local-to-global singular-value calibration used to construct the FLICKER targets, and further control the approximation errors introduced by the sequential projected FLICKER iterations. Combining these components allows us to translate the federated spectral approximation into exact recovery on both sides of the bipartite network.}

\section{Simulation}\label{sec:simu}

\subsection{Data Generation and Comparison Methods}\label{subsec:datageneration}

We consider the bipartite SBM. For any pair of nodes
$i \in \mathcal{L}$ and $j \in \mathcal{R}$, the edge $b_{ij}$ is generated as
$b_{ij} \sim \mathrm{Bernoulli}\big(G(z_1(i), z_2(j))\big)$,
where $z_1(i) \in \{1,\ldots,L_1\}$ denotes the community label of node $i$ on the left
side and $z_2(j) \in \{1,\ldots,L_2\}$ denotes the community label of node $j$ on the
right side. We consider $K$ servers, where server $k$ stores the rows indexed by
$\mathcal{L}^{(k)} \subset \mathcal{L}$ and all columns in $\mathcal{R}$. Our goal is to
recover the community labels for nodes in both $\mathcal{L}$ and $\mathcal{R}$ without
access to the full matrix $B$ at any single location.

We consider nine combinations of the sample size $n$ and the number of servers $K$, with
$n \in \{2000, 4000, 8000\}$ and $ K \in \{5, 10, 20\}$.
We fix $L_1 = 2$ and $L_2 = 3$ communities, and take the block matrix
\[
G
= \frac{1}{50}\begin{pmatrix}
1 + d & 0.5 + d & 0.3 + d\\
0.3+d & 0.5 + d & 1 + d
\end{pmatrix},
\]
where $d$ controls the overall edge density. We consider
$d \in \{-0.2, -0.1, 0, 0.1, 0.2, 0.3\}$.
{Under the balanced community design, the expected edge density is $(0.6+d)/50$, ranging from approximately $0.8\%$ at $d=-0.2$ to $1.8\%$ at $d=0.3$, corresponding to sparse networks. Smaller values of $d$ represent sparser networks.
In this setting, $d$ controls the background density while maintaining the separation between the community connection probabilities.}

We apply spectral community detection on top of several federated PCA/SVD schemes: our method BiFLICKER, DisPCA in \cite{disPCA}, DistPCA in \cite{DistPCA}, FastPCA in \cite{FastPCA}, and the oracle method that has access to the full matrix $B$.
For DisPCA and DistPCA, each server performs PCA locally and sends a small number of
leading eigenvectors (and associated eigenvalues) to the center, which then applies PCA again to the aggregated local information. For all methods, we extract $L=2$ leading singular vectors for downstream clustering.
{For BiFLICKER, we also recover only the leading $L=2$ singular directions, not the full spectrum. Hence, we apply Algorithm~\ref{alg:bi}, with target singular values as rescaled top singular values from one local server. The grid-search procedure in Algorithm~\ref{alg:lambda}, to recover the full spectrum, is not used here.
We set the number of iterations to $T=50K$ in all experiments.}
\subsection{Estimation of $U$ and $z_1$}

We first focus on the left node set $\mathcal{L}$. Recall that $L_1 = 2$, so community
detection on $\mathcal{L}$ is comparatively easier than on $\mathcal{R}$. We examine both
the estimation of the left singular vectors $U$ and the resulting community labels
$z_1$.

\begin{figure}[htbp]
\centering
\includegraphics[width = 0.98\textwidth]{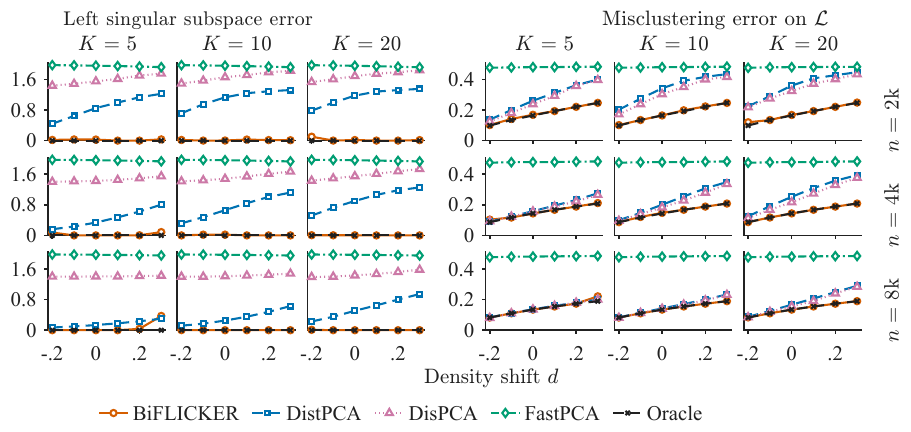}
\caption{Left: estimation errors of the left singular subspace
$\mathrm{Err}(\hat{U};U)$ in \eqref{eqn:errorU}.
Right: misclustering errors on $\mathcal{L}$,
$\mathrm{Err}(\hat{z}_1; z_1)$ in \eqref{eqn:errorz}.
}
\label{fig:left}
\end{figure}

Let $U \in \mathbb{R}^{n\times L}$ be the matrix of the leading $L$ left singular vectors
of the full matrix $B$, and let $\hat{U}$ be the corresponding estimate produced by each
method. We measure the error in estimating the left singular subspace by
\begin{equation}\label{eqn:errorU}
\mathrm{Err}(\hat{U}; U)
= \|\hat{U}\hat{U}^\top - U U^\top\|_F.
\end{equation}
This quantity is zero if and only if the estimated and oracle subspaces coincide. The
results over $100$ repetitions are reported in the left panel of
Figure~\ref{fig:left}.

For community detection on $\mathcal{L}$, we define the misclustering error as the smallest
proportion of misclassified nodes over all permutations $\pi$ of the community labels:
\begin{equation}\label{eqn:errorz}
\mathrm{Err}(\hat{z}_1; z_1)
= \frac{1}{n}\,\min_{\pi:[L_1]\to [L_1]}
\sum_{i=1}^n \mathbf{1}\{\hat{z}_1(i) \neq \pi(z_1(i))\}.
\end{equation}
For each method, we compute $\hat{z}_1$ from the estimated left singular vectors and
compare it with the true labels $z_1$. The results over $100$ repetitions are shown in the
right panel of Figure~\ref{fig:left}.

Across all $(n,K,d)$ configurations, BiFLICKER outperforms the other federated methods
in both subspace estimation and community recovery. DistPCA behaves similarly to
BiFLICKER in the ``large-$n$, small-$K$'' regime, but its performance deteriorates
substantially when $n$ decreases or $K$ increases, reflecting the loss of information at
each local server. A similar degradation is observed for DisPCA. In contrast, BiFLICKER
remains close to the oracle benchmark across all settings, demonstrating its robustness to
both increased federated heterogeneity and smaller local sample sizes.

\subsection{Estimation of $V$ and $z_2$}

We now turn to the right node set $\mathcal{R}$, which has $L_2 = 3$ communities. Note
that the mean matrix $Q = Z_1 G Z_2^\top$ has rank $L = 2$, so we use the first two right
singular vectors as $V$ and then apply $k$-means with $L_2 = 3$ clusters to recover
$z_2$.

\begin{figure}[htbp]
\centering
\includegraphics[width = 0.98\textwidth]{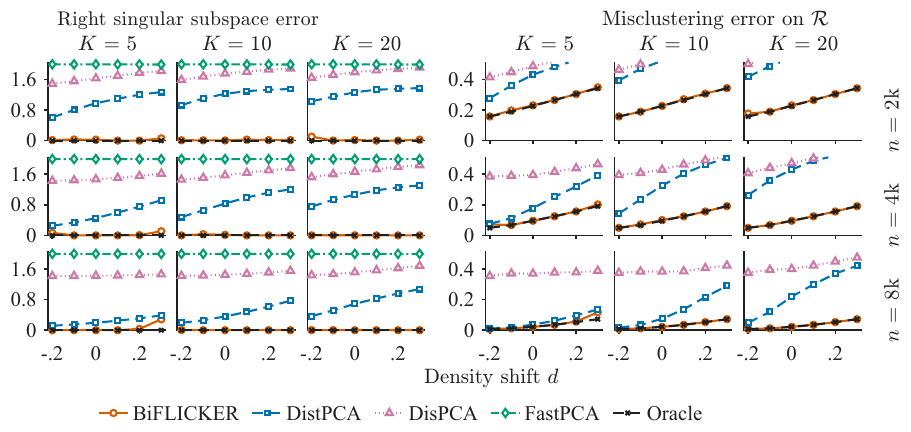}
\caption{Left: estimation errors of the right singular subspace
$\mathrm{Err}(\hat{V};V)$.
Right: misclustering errors on $\mathcal{R}$,
$\mathrm{Err}(\hat{z}_2; z_2)$.}
\label{fig:right}
\end{figure}

We use the same subspace error definition as in \eqref{eqn:errorU}, replacing $U$ by $V$,
and the same misclustering error \eqref{eqn:errorz}, replacing $z_1$ by $z_2$.
The errors over $100$ repetitions are summarized in Figure~\ref{fig:right}.

When clustering $\mathcal{R}$ with $L_2 = 3$ communities, the oracle method itself becomes
slightly more challenging: compared to the $\mathcal{L}$-side results, the error is larger
for small $n=2000$ but decreases again for large $n=8000$. This is typical when the
number of clusters increases.

This additional difficulty is amplified for federated methods. Except for BiFLICKER, all
federated approaches exhibit noticeably higher misclustering errors on $\mathcal{R}$. In
particular, DisPCA attains relatively low error for $\mathcal{L}$ but suffers error rates
of at least $0.4$ on $\mathcal{R}$, even in the most favorable setting
$(n,K) = (8000,5)$. In contrast, BiFLICKER achieves accuracy on $\mathcal{R}$ that is
almost indistinguishable from the oracle across the studied settings.

{Figures~\ref{fig:left} and~\ref{fig:right} also illustrate the effect of network density. As $d$ increases, the common background connectivity increases, without strengthening the contrasts between communities.
The resulting increase in Bernoulli variation makes the community detection problem more challenging.
BiFLICKER nevertheless remains close to the centralized oracle throughout the entire density range, including the low-density settings with $d=-0.2$ and $-0.1$.}

\subsection{Convergence}\label{subsec:conv}

\begin{figure}[htbp]
\centering
\includegraphics[width = 0.9\textwidth]{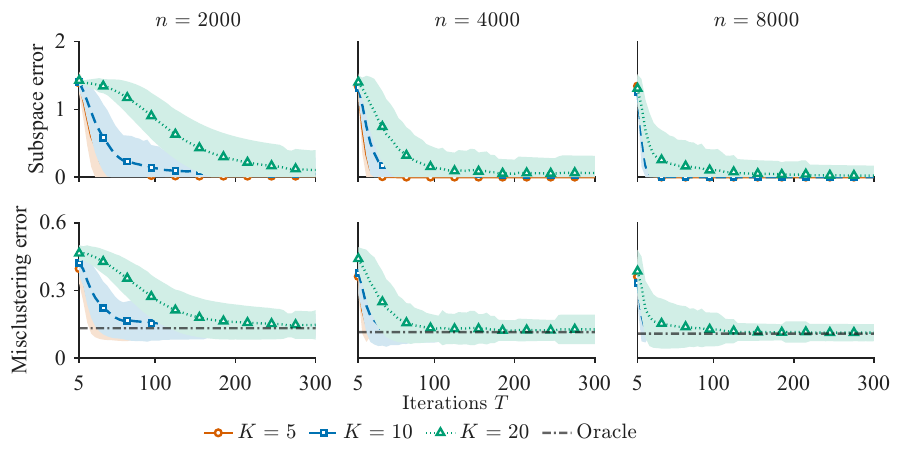}
\caption{Top: estimation errors of the left singular subspace
$\mathrm{Err}(\hat{U};U)$ versus $T$.
Bottom: misclustering errors on $\mathcal{L}$ versus $T$.
Curves show mean error $\pm$ one standard deviation for
$K = 5$ servers (solid), $K = 10$ servers (dashed), and $K = 20$ servers (dotted).}
\label{fig:conv}
\end{figure}

Finally, we examine the convergence of BiFLICKER with respect to the number of iterations
$T$. We use the same data generation scheme as in Section~\ref{subsec:datageneration} with
$d = 0$. For each $n \in \{2000, 4000, 8000\}$ and each
$K \in \{5, 10, 20\}$, we run BiFLICKER with
$T \in \{5, 10, 15, \ldots, 300\}$,
and record the error in estimating the left singular subspace $U$ and the misclustering
error on $\mathcal{L}$ over $100$ repetitions. The results are shown in
Figure~\ref{fig:conv}. Results for the right singular vectors $V$ and node set $\mathcal{R}$ are very similar and omitted here.

Figure~\ref{fig:conv} shows that, as $n$ increases, BiFLICKER requires fewer iterations to
reach a given accuracy level. For fixed $n$, increasing the number of servers $K$ reduces
the sample size per server and consequently requires more iterations for convergence.
Overall, even in the most challenging case $(n,K) = (2000,20)$, setting $T = 300$ is
sufficient to achieve near-oracle performance in most replications.

\section{LastFM Data Analysis}\label{sec:data}
We analyze the LastFM listening-history data set, available from the original data repository at\footnote{\url{https://ocelma.net/MusicRecommendationDataset/lastfm-1K.html}}~\citep{celma2010music}. Let $\mathcal{L}$ denote the set of users and $\mathcal{R}$ the set of artists. For a user $i\in\mathcal{L}$ and artist $j\in\mathcal{R}$, we set $B(i,j)=1$ if user $i$ listened to artist $j$ at least once, and $B(i,j)=0$ otherwise. The resulting binary user--artist network is used for spectral estimation and community detection. It is available at \url{https://github.com/wanjiew/BiFLICKER}.

The pre-processing is as follows. We retain countries with at least $30$ users and artists with valid identifiers. Then iteratively remove users connected to fewer than $50$ retained artists and artists connected to fewer than $10$ users or at least $40\%$ of the retained users. The upper-degree filter excludes extremely popular artists that provide little information for distinguishing user preferences. The final network contains $n=511$ users from $7$ countries, $m=8{,}632$ artists, and $288{,}628$ edges, corresponding to an edge density of $6.54\%$.

The country information provides a natural way to represent geographically separated user populations. In many online platforms, user interaction data may be maintained by regional services, while users across regions interact with a largely common set of items. Motivated by this setting, we treat each country as one local server. Specifically, server $k$ contains the users from country $k$ and stores the corresponding submatrix $B_k\in\{0,1\}^{n_k\times m}$, while the artist set is shared across all servers. Hence, $B=[B_1^\top,\ldots,B_7^\top]^\top$. Table~\ref{tab:lastfm_server} summarizes the local datasets, where the sizes and average degrees are substantially heterogeneous across these local servers. In the federated analysis, the local matrices $B_k$ remain on the corresponding servers; only the matrix--vector products required by FLICKER are communicated.

\begin{table}[htbp]
\centering
\caption{Summary of the 7 country servers in the cleaned LastFM user--artist network.}
\label{tab:lastfm_server}
\small
\setlength{\tabcolsep}{4pt}
\begin{tabular}{lccccccc}
\toprule
& US
& UK
& Poland
& Germany
& Finland
& Norway
& Canada \\
\midrule
Users $n_k$
& 216 & 121 & 46 & 36 & 31 & 31 & 30 \\
Mean user degree
& 575 & 636 & 487 & 560 & 405 & 493 & 569 \\
Density
& 6.7\% & 7.4\% & 5.6\% & 6.5\% & 4.7\% & 5.7\% & 6.6\% \\
\bottomrule
\end{tabular}
\end{table}

We take $L = 7$ left singular vectors and \blue{use $L_1 = 8$ user communities} for all analyses below. Here, $L = 7$ is suggested by the spectral diagnostics. \blue{$L_1 = 8$ is selected by maximizing the average silhouette score among $7 \leq L_1 \leq 30$ communities, under the restriction the minimum community size is no smaller than 10.}

\begin{figure}[htbp]
\centering
  \includegraphics[width=0.9\linewidth]{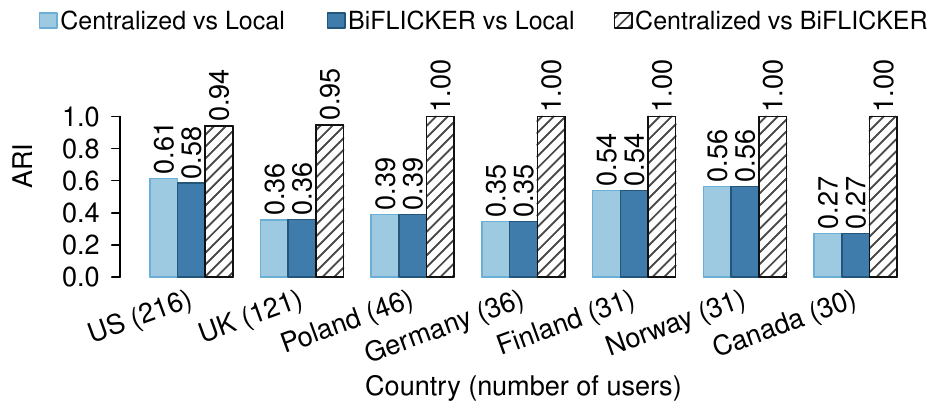}
\caption{Agreement of user-community assignments within each country. The three bars
compare the centralized, local, and Sequential BiFLICKER assignments using ARI.}
\label{fig:lastfm-ari}
\end{figure}

We first compare three community detection procedures: local, central, and federated. In the local analysis, we only consider the local data $B_k$ to find the communities. The centralized analysis applies the same community detection procedure to the complete user--artist matrix $B$, which serves as a benchmark. In the federated analysis, we apply Sequential BiFLICKER, which estimates a common global embedding while the country-specific matrices remain local.
Figure~\ref{fig:lastfm-ari} compares the Adjusted Rand Index (ARI, \cite{HubertArabie1985}) between the three community estimation methods within each country. The centralized and local assignments have relatively low agreement, with ARIs ranging from $0.27$ to $0.61$, showing that the information available on an individual server is generally insufficient to reproduce the global community structure. US, the server with the most users, shows the highest agreement with central analysis, which further emphasizes this observation. The Sequential BiFLICKER--local comparison gives a similar pattern. In contrast, the centralized and Sequential BiFLICKER assignments agree closely across all 7 countries, with ARIs between \blue{$0.94$} and $1.00$ and an overall ARI of \blue{$0.96$}. Thus, Sequential BiFLICKER recovers essentially the same global structure as the centralized analysis while keeping the country-specific matrices local. We therefore base the following community interpretation on the Sequential BiFLICKER partition.

For Sequential BiFLICKER, we use the US server to obtain top 7 local singular values, and then rescale them by $\sqrt{511/216}$ to form the target values. \blue{The step size for component $j$ is $\eta_j=\widetilde{\sigma}_j^{-4}$. We stop the iteration when the relative change is lower than a $5\times 10^{-4}$ for two consecutive checks. Sequential BiFLICKER requires 3 updates for the first singular vector and 78 updates for all 7 vectors. Therefore, the communication cost for the leading singular vector is about 1/5 of transferring the whole matrix. Detailed implementation settings, communication cost, and selection of $L_1$ are provided in the Supplementary Material.}

\begin{figure}[!htbp]
\centering
  \includegraphics[width=0.95\linewidth]{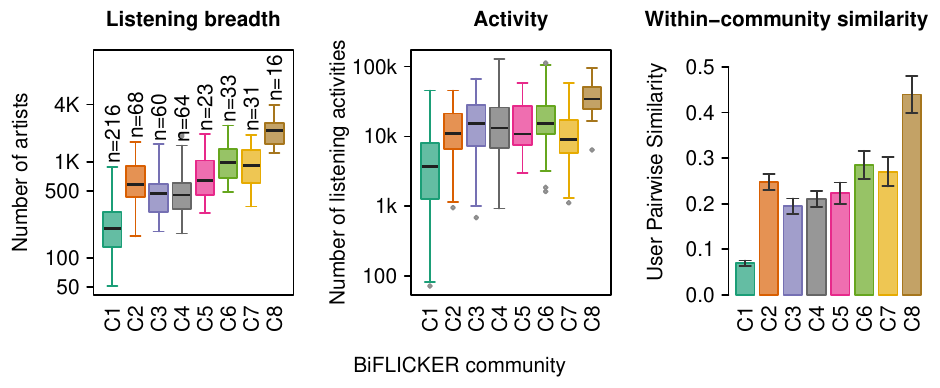}
\caption{Listening behavior by Sequential BiFLICKER user community, summarizing listening breadth (left panel), activity (middle panel), and within-community similarity (right panel) with $95\%$ leave-one-user-out jackknife
confidence intervals.}
\label{fig:lastfm-behavior}
\end{figure}

We then examine the $L_1=8$ user communities obtained by Sequential BiFLICKER.
We first investigate their overall listening behavior using listening breadth, activities, and within-community similarities. Listening breadth is the
number of artists heard by a user (i.e., degree), while activity is the total number of listening events.
We use the average pairwise cosine similarity between users' artist-incidence vectors within a community to evaluate the coherence. Results are summarized in Figure~\ref{fig:lastfm-behavior}.

The communities exhibit clear differences along these behavioral dimensions.
Community~1, the largest group, has the lowest
breadth, activity, and within-community similarity. It therefore represents a relatively low-engagement and heterogeneous group of listeners.
Community~8 is at the opposite extreme: it has the greatest breadth and activity, together with the strongest overlap in users' artist sets.
This pattern is characteristic of avid listeners with broad but relatively cohesive musical interests.
Communities~2--7 are all relatively active, but they differ markedly in listening breadth and within-community similarity. Communities~3--5 have more moderate breadth and lower overlap in their artist sets, whereas Communities~6 and~7 listen to a broader range of artists and exhibit stronger within-community overlap. Community~2 lies between these two patterns, with intermediate breadth and similarity.

Table~\ref{tab:lastfm-profile} further characterizes the communities through their representative artists.
Communities~3, ~4, and ~5 have metal, emo/pop-punk, and hip hop/R\&B profiles, respectively. Within these broad genres, individual users appear to concentrate on relatively small sets of preferred artists and substyles, resulting in lower listening breadth. As these preferred sets differ across users, the within-community pairwise similarity also remains relatively low.
Communities~6 and~7 instead combine broad listening with more strongly shared repertoires, represented by classic pop rock and soul, and by downtempo and electronica, respectively. Community~2 lies between these patterns with an indie rock and post-punk profile, while Community~8 has a roots-oriented profile spanning country, blues, folk, and psychedelic rock. Community~1 has no comparably distinctive artist core, consistent with its low within-community similarity.

\begin{table}[!htbp]
\centering
\caption{Country composition and representative artists of the Sequential BiFLICKER user communities. Country percentages are relative to community size. }
\label{tab:lastfm-profile}
\small
\setlength{\tabcolsep}{4pt}
\begin{tabular}{@{}cc>{\footnotesize}l>{\scriptsize}l@{}}
\toprule
Community & Users & \multicolumn{1}{l}{Main countries}
& \multicolumn{1}{l}{Representative artists} \\
\midrule
C1 & 216 & US 41.7\%; UK 21.8\%
& Aerosmith; Mgmt; Lez Zeppelin \\

C2 & 68 & US 42.6\%; UK 29.4\%
& The Go! Team; British Sea Power; Art Brut \\

C3 & 60 & Finland 28.3\%; US 26.7\%
& Pantera; Megadeth; Disturbed \\

C4 & 64 & US 70.3\%; UK 14.1\%
& Taking Back Sunday; Fall Out Boy; Paramore \\

C5 & 23 & US 52.2\%; UK 17.4\%
& Missy Elliott; Ludacris; Jay-Z \\

C6 & 33 & UK 48.5\%; US 30.3\%
& Dexys Midnight Runners; Rod Stewart; Squeeze \\

C7 & 31 & UK 35.5\%; Poland 32.3\%
& The Cinematic Orchestra; Tosca; Nightmares On Wax \\

C8 & 16 & US 62.5\%; UK 31.2\%
& The Flying Burrito Brothers; Howlin' Wolf; Hank Williams, Jr. \\
\bottomrule
\end{tabular}
\end{table}
The country composition in Table~\ref{tab:lastfm-profile} also shows that the recovered communities do not simply reproduce the country-based server partition.
Each of the larger communities contains users from multiple countries, although some geographic differences remain: for example, Community~4 is more strongly represented by US users, Community~6 by UK users, and Community~7 contains substantial proportions of both UK and Polish users.
Thus, the global communities recovered by Sequential BiFLICKER cut across the local servers while exhibiting distinct listening behaviors and artist preferences.

\section*{Acknowledgments}

The author thanks the Editor, the Associate Editor, and the
three anonymous reviewers for their constructive comments and suggestions, which have improved the manuscript.

\section*{Funding}

This work was supported by the Singapore Ministry of Education
under grants Tier-1-A-8001451-00-00 and A-8003581-00-00.

\section*{Declaration of generative AI use}

The author used OpenAI Codex (GPT-6 Astra) for code revision
and language refinement. The author verified the originality and accuracy of the AI-assisted content and takes full responsibility for the entire manuscript.

\section*{Disclosure Statement}

The author reports there are no competing interests to declare.

\clearpage
\section*{Supplementary Material}
\phantomsection\label{supplementary-material}
\addcontentsline{toc}{section}{Supplementary Material}
\setcounter{section}{0}
\renewcommand{\thesection}{S\arabic{section}}
\providecommand{\theHsection}{}
\renewcommand{\theHsection}{supp.\arabic{section}}
\setcounter{equation}{0}
\renewcommand{\theequation}{S\arabic{equation}}
\providecommand{\theHequation}{}
\renewcommand{\theHequation}{supp.\arabic{equation}}
\setcounter{figure}{0}
\renewcommand{\thefigure}{S\arabic{figure}}
\providecommand{\theHfigure}{}
\renewcommand{\theHfigure}{supp.\arabic{figure}}
\setcounter{table}{0}
\renewcommand{\thetable}{S\arabic{table}}
\providecommand{\theHtable}{}
\renewcommand{\theHtable}{supp.\arabic{table}}
\setcounter{theorem}{0}
\renewcommand{\thetheorem}{S\arabic{theorem}}
\providecommand{\theHtheorem}{}
\renewcommand{\theHtheorem}{supp.\arabic{theorem}}
\setcounter{assumption}{0}
\renewcommand{\theassumption}{S\arabic{assumption}}
\providecommand{\theHassumption}{}
\renewcommand{\theHassumption}{supp.\arabic{assumption}}
\setcounter{corollary}{0}
\renewcommand{\thecorollary}{S\arabic{corollary}}
\providecommand{\theHcorollary}{}
\renewcommand{\theHcorollary}{supp.\arabic{corollary}}
\setcounter{prop}{0}
\renewcommand{\theprop}{S\arabic{prop}}
\providecommand{\theHprop}{}
\renewcommand{\theHprop}{supp.\arabic{prop}}
\setcounter{pro}{0}
\renewcommand{\thepro}{S\arabic{pro}}
\providecommand{\theHpro}{}
\renewcommand{\theHpro}{supp.\arabic{pro}}
\setcounter{definition}{0}
\renewcommand{\thedefinition}{S\arabic{definition}}
\providecommand{\theHdefinition}{}
\renewcommand{\theHdefinition}{supp.\arabic{definition}}
\setcounter{example}{0}
\renewcommand{\theexample}{S\arabic{example}}
\providecommand{\theHexample}{}
\renewcommand{\theHexample}{supp.\arabic{example}}
\setcounter{lemma}{0}
\renewcommand{\thelemma}{S\arabic{lemma}}
\providecommand{\theHlemma}{}
\renewcommand{\theHlemma}{supp.\arabic{lemma}}
\setcounter{proposition}{0}
\renewcommand{\theproposition}{S\arabic{proposition}}
\providecommand{\theHproposition}{}
\renewcommand{\theHproposition}{supp.\arabic{proposition}}
\setcounter{remark}{0}
\renewcommand{\theremark}{S\arabic{remark}}
\providecommand{\theHremark}{}
\renewcommand{\theHremark}{supp.\arabic{remark}}
\setcounter{algorithm}{0}
\renewcommand{\thealgorithm}{S\arabic{algorithm}}
\providecommand{\theHalgorithm}{}
\renewcommand{\theHalgorithm}{supp.\arabic{algorithm}}

This supplementary file contains the implementation details and additional results about the LastFM data set in main manuscript, an additional movie data analysis, and the full proof of the strong consistency result for BiFLICKER. Section~\ref{app:lastfm} introduces the implementation details, selection of $L$ and $L_1$, and community interpretation details about LastFM dataset. Section~\ref{app:movie} applies parallel BiFLICKER on a Douban movie dataset. Section~\ref{sec:flickerconv} provides the proof for lemmas, propositions, and theorems about the FLICKER algorithm. In Sections~\ref{sec:aux-lemmas}--\ref{sec:strong-consistency}, we prove the strong consistency theorem about BiFLICKER under the sparse bipartite SBM.

\section{LastFM Data Analysis: Details}\label{app:lastfm}

\subsection{Data Preprocessing}

We provide additional details on the preprocessing of the LastFM data used in Section~\ref{sec:data}. We first retain countries represented by at least $30$ users and remove artists without valid MusicBrainz identifiers (MBIDs).
We then construct the binary user--artist network and iteratively apply degree filtering until the retained bipartite network no longer changes. Specifically, a user is retained if connected to at least $50$ retained artists, while an artist is retained if connected to at least $10$ but fewer than $40\%$ of the retained users. The upper-degree restriction removes ubiquitous artists that provide little information for distinguishing user preferences.

\begin{figure}[!htbp]
    \centering
    \includegraphics[width=.82\textwidth]{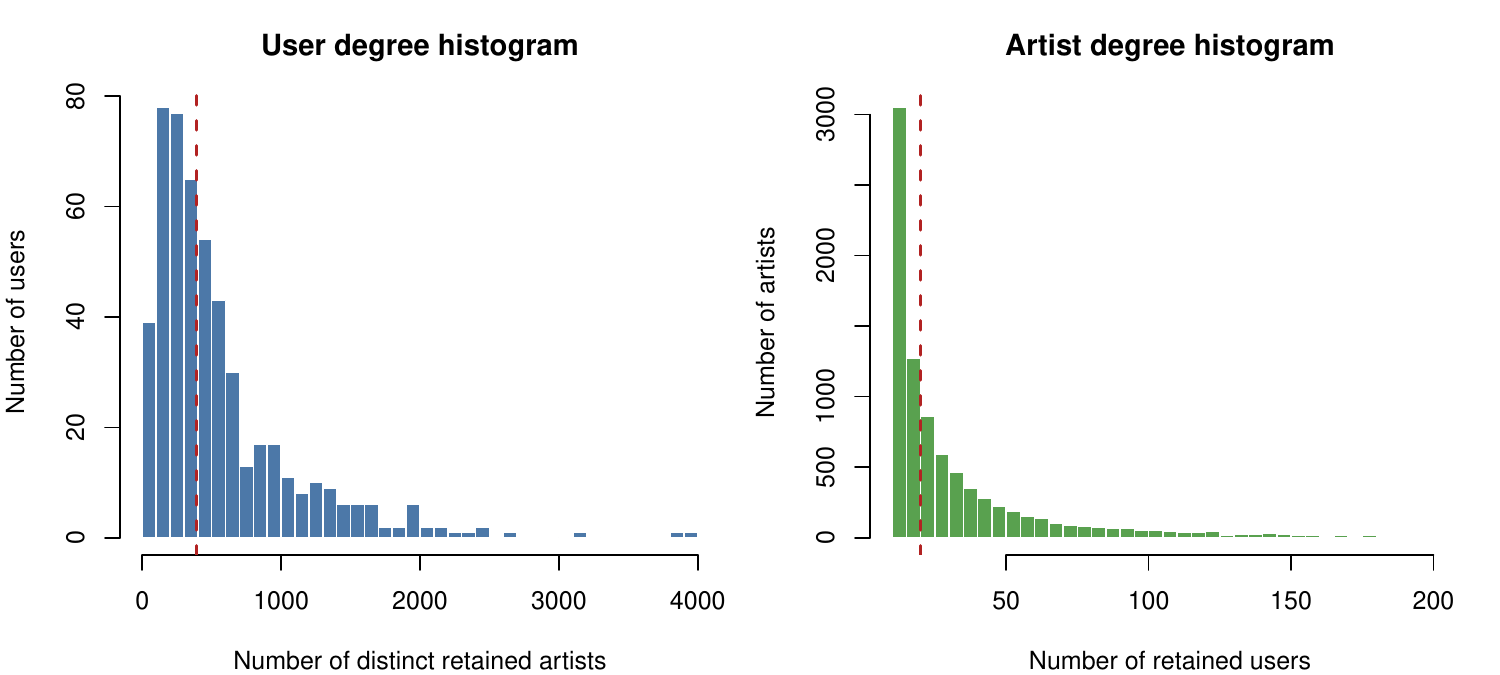}
    \caption{Degree distributions in the retained LastFM user--artist network.}
    \label{fig:lastfm-degree}
\end{figure}

There are $539$ users from the seven retained countries before the iterative degree filtering and $511$ users afterward. The final network contains $8{,}632$ artists and $288{,}628$ binary user--artist edges. Figure~\ref{fig:lastfm-degree} shows the degree distributions in the resulting network. User degrees remain highly heterogeneous, with median $390$ and mean $564.8$, whereas the artist degrees are more concentrated after filtering, with median $20$ and mean
$33.4$.

\subsection{Selection of the Number of Embeddings}

\begin{figure}[htbp]
    \centering
    \includegraphics[width=.55\textwidth]{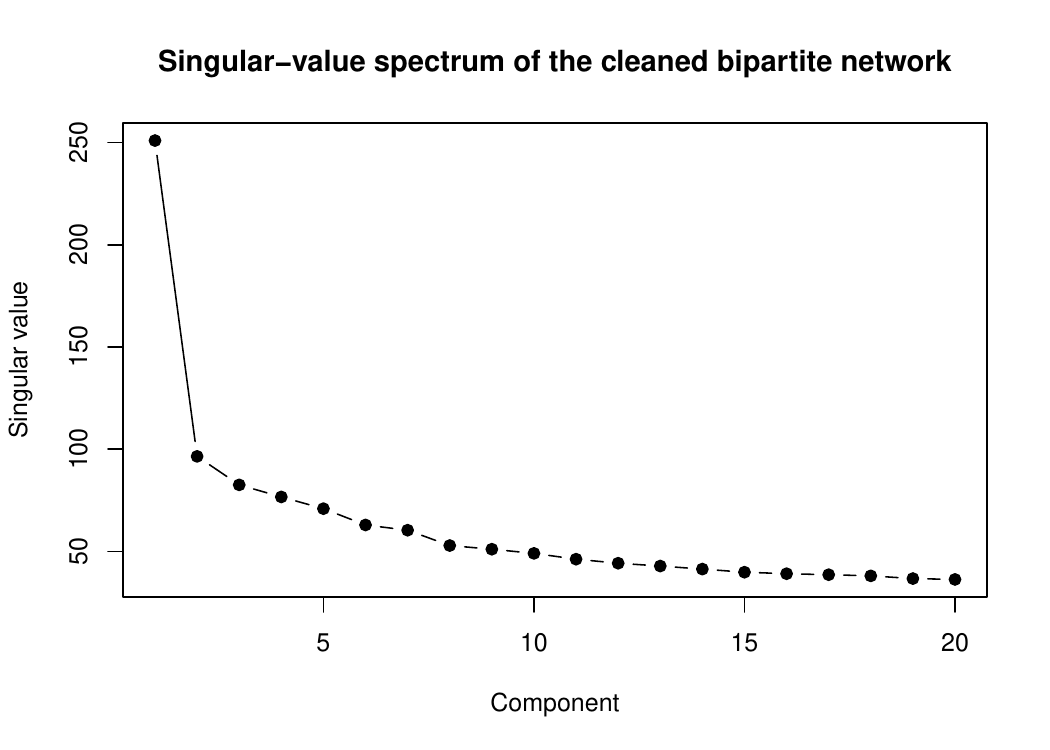}
    \caption{Leading singular values of the retained LastFM user--artist
    network.}
    \label{fig:lastfm-spectrum}
\end{figure}

We use the centralized singular-value spectrum as a diagnostic for the dimension of the user embedding. Figure~\ref{fig:lastfm-spectrum} shows the first $20$ singular values of the retained user--artist matrix. The spectrum decays gradually after the dominant first component, with additional local separations around components $5$ and $7$. In particular, $\sigma_5/\sigma_6=1.127$ and $\sigma_7/\sigma_8=1.142$. We use $L=7$ in the main analysis to obtain the finer of these two resolutions.

\subsection{Selection of the Number of Communities}
Now we consider the selection of $L_1$ and $L_2$, yielding $\min\{L_1, L_2\} = L$. We fix $L = 7$ to find out the embeddings, and discuss the best partition on this embedding.

Let $\widehat U\in\mathbb R^{n\times7}$ be the final user coordinates estimated
by the left-side Sequential BiFLICKER iteration. Artist coordinates are
constructed as
\begin{equation}
\widehat V=B^\top\widehat U\,
\operatorname{diag}(\widehat s_1^{-1},\ldots,\widehat s_7^{-1}),
\label{eq:supp-lastfm-right}
\end{equation}
where $\widehat s_j=(\widehat u_j^\top BB^\top\widehat u_j)^{1/2}$.
These are approximate right-side coordinates derived from the same user
estimate; no separate right-side iteration is performed. We apply
$k$-means to the rows of $\widehat U$ and $\widehat V$ using their original
coordinates, without additional row normalization or singular-value
weighting.

For each side and each $k\in\{7,8,\ldots,30\}$, we retain the $k$-means solution with the smallest within-cluster sum of squares among 1,000 random
starts. We then evaluate the resulting
partition using the average silhouette coefficient \citep{Rousseeuw1987}.
For a node $i$, let $a_i(k)$ be its average Euclidean distance to the other
nodes in its own cluster, and let $b_i(k)$ be the smallest of its average
distances to the nodes in each other cluster. Its silhouette is
\begin{equation}
s_i(k)=\frac{b_i(k)-a_i(k)}{\max\{a_i(k),b_i(k)\}},
\qquad S(k)=\frac{1}{N}\sum_{i=1}^{N}s_i(k),
\label{eq:supp-lastfm-silhouette}
\end{equation}
where $N=n$ on the user side and $N=m$ on the artist side. A singleton is assigned silhouette zero.

To avoid choosing small communities with limited interpretability, we only consider partitions with all community sizes no smaller than  $10$. Hence, the selected $L_q$ is
\begin{equation}
\widehat L_q=\underset{\substack{k\in\{7,\ldots,30\}:\\
\min_{1\leq a\leq k}|\widehat C_{q,a}(k)|\geq10}}{\operatorname{arg\,max}}
S_q(k), \quad q = 1, 2,
\label{eq:supp-lastfm-choice}
\end{equation}
with the smaller count preferred for an exact tie.

\begin{figure}[htbp]
\centering
\includegraphics[width=\linewidth]{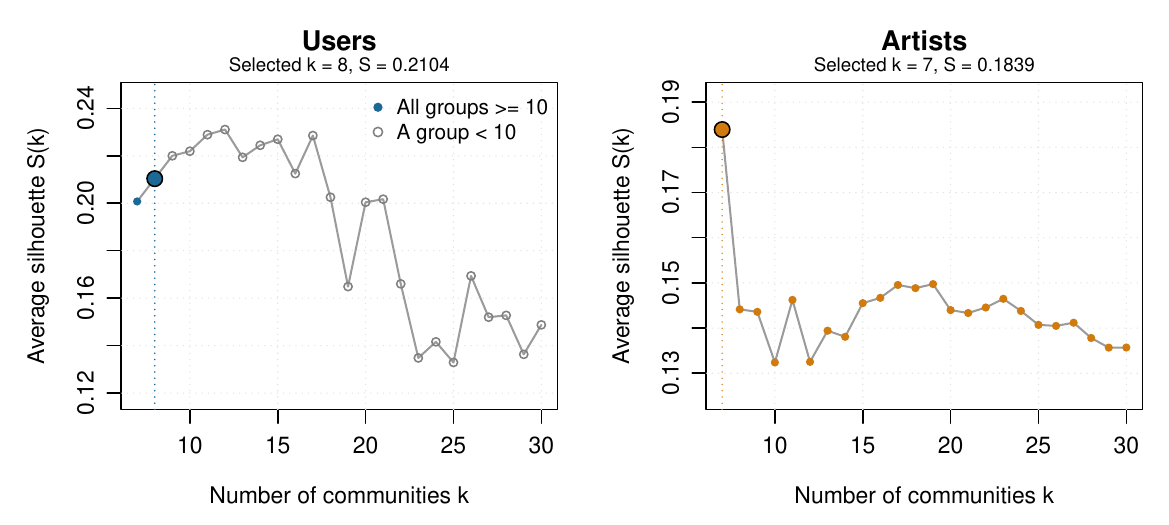}
\caption{Average silhouette coefficients in the estimated
7-dim coordinates.}
\label{fig:supp-lastfm-selection}
\end{figure}

The scores $S(k)$ for $7 \leq k \leq 30$ are summarized in Figure \ref{fig:supp-lastfm-selection}, with users in the left panel and artists in the right panel. Under the restriction that the minimum community size is at least 10, we choose $L_1 = 8$ and $L_2 = 7$, which yields the largest average silhouettes scores. Their smallest groups contain 16 users and 314 artists, respectively.
Without the restriction, $L_2 = 7$ largely outperforms other choices, and $L_1$ is preferred to be 12 with a very similar score. Hence, our choices are reasonable. Furthermore, $L = \min\{L_1, L_2\} = 7$.

This is an internal clustering heuristic conditional on the fitted embeddings. It favors compact and separated communities. However, it can overlook smaller communities, and does not quantify uncertainty in the spectral embedding. We use it to give a heuristic selection of $L_1$ and $L_2$.

\subsection{Implementation Details of Sequential BiFLICKER}

The United States is the largest local server, containing $216$ of the $511$ retained users, and is used to provide the reference spectral information.
Let $\sigma_i^{(\mathrm{US})}$ denote its $i$th local singular value. For $i=1,\ldots,7$, Sequential BiFLICKER uses
\[
    \widetilde{\sigma}_i
    =
    \sqrt{\frac{511}{216}}\,
    \sigma_i^{(\mathrm{US})}
\]
as the target singular value.
The corresponding local left singular direction is mapped to a shared right-side seed, which initializes the distributed left coordinates. The targets are processed in decreasing order. Only the
distributed left coordinates are iterated; when recovering direction $i$,
they are projected onto the orthogonal complement of the first $i-1$
recovered left directions. No separate right-side gradient iteration is
performed.

For target $\widetilde{\sigma}_i$, the step size is
\[
    \eta_i=\widetilde{\sigma}_i^{-4}.
\]
Each direction is given a maximum of
$\lceil 10\log(511)\rceil=63$ updates. The ordinary relative change between successive left Rayleigh-based singular-value estimates is checked after every update. The first update establishes the comparison baseline, and there is no minimum iteration requirement. Early stopping occurs when this relative change is at most $5\times10^{-4}$ for two consecutive checks. This is an empirical stability criterion rather than a singular-vector error
certificate.

The seven components stop after $3,7,11,29,11,8,$ and $9$ updates,
respectively, for $78$ updates in total. All seven satisfy the early-stopping criterion, and none reaches the update cap.
Estimating all 7 singular vectors requires approximately $162.50$ MB of communication, including initialization, iterative updates, and convergence checks with intermediate products reused. By comparison, transferring the complete user--artist matrix once requires $35.29$ MB. Thus, the total communication for spectral estimation is approximately $4.60$ times that of a single matrix transfer. However, if we are interested in the first singular vector only, then the communication cost is $7.37$ MB, about 1/5 of the whole matrix transfer cost.

Table~\ref{tab:lastfm-spectrum-comparison} compares the resulting singular-value
estimates with those obtained from the centralized SVD. The two spectra are
very close: the largest absolute relative difference is $0.97\%$ at component
$5$, followed by $0.51\%$ at component $4$.

\begin{table}[htbp]
\centering
\small
\setlength{\tabcolsep}{5pt}
\begin{tabular}{crrrr}
\toprule
Component &
Target $\widetilde{\sigma}_i$ &
Centralized &
Sequential BiFLICKER &
Relative difference \\
\midrule
1 & 269.875 & 250.991 & 250.991 &  $0.00\%$ \\
2 & 102.865 &  96.448 &  96.403 & $-0.05\%$ \\
3 &  92.053 &  82.502 &  82.229 & $-0.33\%$ \\
4 &  77.837 &  76.594 &  76.203 & $-0.51\%$ \\
5 &  75.150 &  70.858 &  71.545 &  $0.97\%$ \\
6 &  71.256 &  62.878 &  62.887 &  $0.01\%$ \\
7 &  69.376 &  60.315 &  60.382 &  $0.11\%$ \\
\bottomrule
\end{tabular}
\caption{Comparison of the target, centralized, and Sequential BiFLICKER
singular values for the first seven components. Relative differences are
computed relative to the centralized singular values.}
\label{tab:lastfm-spectrum-comparison}
\end{table}

\subsection{Representative Artists}

We describe how the representative artists used to interpret the user
communities in Section~\ref{sec:data} are selected, and extend the
three-artist summaries in the main text to five artists per community.
We also report their coverage, defined as the proportion of users in each
community who listened to at least one of the five listed artists.

The selection is based on binary user--artist incidence, consistent with
the network used for community detection. For community $c$, an artist is
considered as a distinctive candidate if it is heard by at least 5 users and at least $10\%$ of the users in that community, and its prevalence
within the community exceeds its prevalence among users outside the
community. Candidate artists are ranked by the difference between
within- and outside-community prevalence. If there are ties, we use a smoothed log-odds ratio for a finer ranking.

The 5 highest-ranked artists are retained as representative artists. If
fewer than five positively enriched artists satisfy the criteria, we use the most prevalent artists satisfying the same support requirement as a fallback.
Actually, this fallback is required only for Community~1, for which no comparably distinctive positively enriched artist core is identified. Table~\ref{tab:lastfm-artists-full} reports the full five-artist summaries. The
top-five artists cover nearly all users in Communities~2--$8$, whereas the
coverage is lower for the more heterogeneous Community~1.

\begin{table}[htbp]
\centering
\small
\setlength{\tabcolsep}{4pt}
\begin{tabular}{crp{9.1cm}c}
\toprule
Community & Users & Representative artists & Coverage \\
\midrule
C1 & 216 &
Aerosmith; Mgmt; Lez Zeppelin; John Lennon; Morrissey
& 70.8\% \\

C2 & 68 &
The Go! Team; British Sea Power; Art Brut; The Futureheads; The Rakes
& 95.6\% \\

C3 & 60 &
Pantera; Megadeth; Disturbed; Type O Negative; Rammstein
& 98.3\% \\

C4 & 64 &
Taking Back Sunday; Fall Out Boy; Paramore; Hellogoodbye;
Panic At The Disco
& 96.9\% \\

C5 & 23 &
Missy Elliott; Ludacris; Jay-Z; LL Cool J; Fugees
& 100.0\% \\

C6 & 33 &
Dexys Midnight Runners; Rod Stewart; Squeeze; The Hollies;
Dusty Springfield
& 100.0\% \\

C7 & 31 &
The Cinematic Orchestra; Tosca; Nightmares On Wax; Jazzanova; Bent
& 100.0\% \\

C8 & 16 &
The Flying Burrito Brothers; Howlin' Wolf; Hank Williams, Jr.;
Os Mutantes; John Fahey
& 100.0\% \\
\bottomrule
\end{tabular}
\caption{Representative artists for the eight Sequential BiFLICKER user
communities. Coverage is the percentage of users in a community who
listened to at least one of the five listed artists. For Community~1,
the listed artists are selected by the prevalence-based fallback
described in the text.}
\label{tab:lastfm-artists-full}
\end{table}

\section{Movie Data Analysis}\label{app:movie}

We analyze a user--movie bipartite network constructed from movie ratings on
\url{douban.com} between April 1, 2023 and June 23, 2023. Let
$\mathcal{L}$ denote the set of users and $\mathcal{R}$ the set of movies. For any
user $i \in \mathcal{L}$ and movie $j \in \mathcal{R}$, we set $B(i,j) = 1$ if user $i$ rated movie $j$ during this period and $B(i,j) = 0$ otherwise.

To mimic a federated setting, we construct five datasets with different user sets but the
same movie set. For dataset $k$, let $n_k$ be the number of users and $m$ be the number
of movies (common across all datasets). Table~\ref{tab:data} reports, for each dataset,
the number of users $n_k$, the common number of movies $m$, and the average number of
movies rated per user. The average number of ratings decreases sharply from dataset~1 to
dataset~5, indicating substantial heterogeneity in user activity across datasets. This
heterogeneity motivates the use of federated community detection to jointly leverage all
datasets.

\begin{table}[htbp]
    \centering
    \begin{tabular}{c|ccccc}
        Dataset & 1 &  2 & 3 & 4 & 5\\
        \hline
        $n_k$ & 75 & 101 & 494 & 1275 & 249\\
        \hline
        $m$ & 3129 & 3129 & 3129 & 3129 & 3129 \\
        \hline
        Average \# of Movies Rated & 774 & 364 & 204 & 95 & 37\\
        \hline
    \end{tabular}
    \caption{Number of users and movies in each dataset, and the average number of movies
    rated per user.}
    \label{tab:data}
\end{table}

Let $B_k \in \{0,1\}^{n_k \times m}$ denote the user-movie adjacency matrix for
dataset $k$, $k=1,\ldots,5$. We treat the five datasets as five federated servers. We
apply BiFLICKER with $T = 5000$ iterations to the collection $\{B_k\}_{k=1}^5$ and
compute the top $10$ left and right singular vectors. Based on the empirical singular
values, we retain the top $6$ components on each side and set $L_1 = L_2 = 6$ for
subsequent clustering. Additional implementation details are provided in the supplementary materials.

We first examine the user communities. Table~\ref{tab:comm} reports the estimated
community sizes for the $L_1 = 6$ user communities in each dataset, together with proportions on server $k$. As the average number of rated movies decreases, the proportion of users in
community~1 increases markedly and eventually dominates dataset~5. This community can be
interpreted as a group of ``general users'' without strong or specific genre preferences.
They are present in all datasets, but become the overwhelming majority in dataset~5, where
each user rates only a small number of movies.

\begin{table}[!htbp]
    \centering
\scalebox{0.7}{
    \begin{tabular}{c|cccccc|c}
        User Community & 1 & 2 & 3 & 4 & 5 & 6 & Total\\
        \hline
        Dataset 1
        & 25 (33.3\%) & 6 (8.0\%) & 10 (13.3\%) & 6 (8.0\%) & 17 (22.7\%) & 10 (13.3\%) & 75\\
        Dataset 2
        & 35 (34.7\%) & 14 (13.9\%) & 18 (17.8\%) & 8 (7.9\%) & 14 (13.9\%) & 11 (10.9\%) & 101\\
        Dataset 3
        & 179 (36.2\%) & 77 (15.6\%) & 85 (17.2\%) & 62 (12.6\%) & 52 (10.5\%) & 38 (7.7\%) & 494\\
        Dataset 4
        & 681 (53.4\%) & 182 (14.3\%) & 116 (9.1\%) & 114 (8.9\%) & 100 (7.8\%) & 81 (6.4\%) & 1275\\
        Dataset 5
        & 217 (87.1\%) & 11 (4.4\%) & 4 (1.6\%) & 5 (2.0\%) & 2 (0.8\%) & 9 (3.6\%) & 249\\
        \hline
        \hline
        Community Size
        & 1142 (52.1\%) & 290 (13.2\%) & 233 (10.6\%) & 195 (8.9\%) & 185 (8.4\%) & 149 (6.8\%) & 2194\\
        \hline
    \end{tabular}}
    \caption{Number of users in each estimated community. Row-wise proportions are shown
    in brackets.}
    \label{tab:comm}
\end{table}

\begin{figure}[!htbp]
    \centering
    \includegraphics[width=0.65\linewidth]{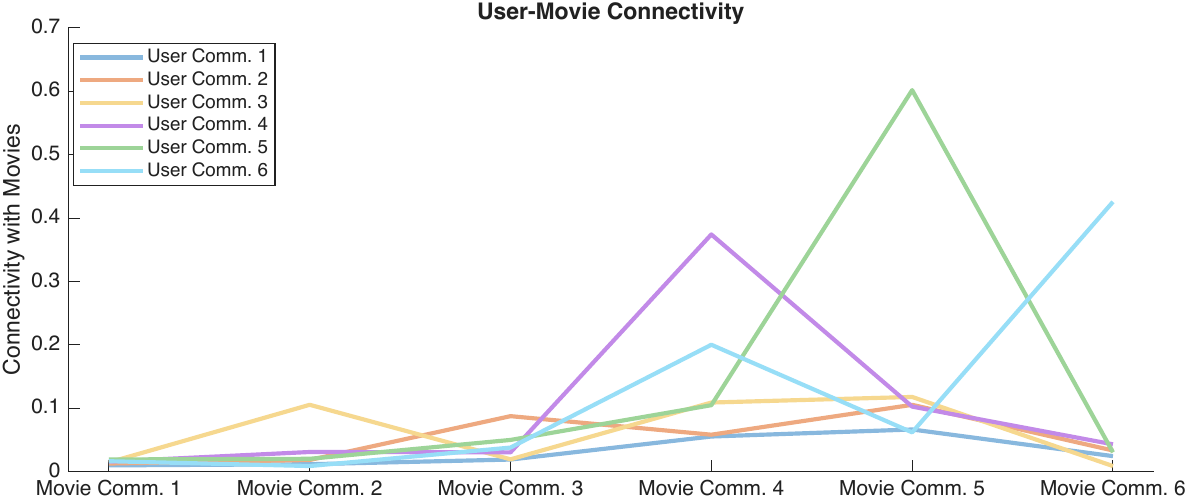}
    \caption{Edge densities between user communities and movie communities.}
    \label{fig:connectivity}
\end{figure}

Next, we investigate how user communities interact with movie communities. Let
$\hat{z}_1$ and $\hat{z}_2$ denote the estimated community label vectors for users and
movies, respectively. For $l_1 \in [L_1]$ and $l_2 \in [L_2]$, define the edge density
between user community $l_1$ and movie community $l_2$ as
\[
d(l_1,l_2)
= \frac{\#\{(i,j): \hat{z}_1(i)=l_1,\ \hat{z}_2(j)=l_2,\ B(i,j)=1\}}
{\#\{(i,j): \hat{z}_1(i)=l_1,\ \hat{z}_2(j)=l_2\}}.
\]
Figure~\ref{fig:connectivity} displays the matrix of densities $d(l_1,l_2)$. The
``general user'' community~1 exhibits relatively low and fairly uniform densities across
all movie communities, consistent with a broad but shallow interest profile. In contrast,
the other user communities display pronounced preferences. User communities 4, 5, and 6
have particularly high densities with movie communities 4, 5, and 6, respectively,
indicating strong one-to-one preference patterns. User community 2 is more strongly
connected to movie communities 3 and 5, while user community 3 shows relatively higher
densities with movie communities 2, 4, and 5, revealing cross-community interaction patterns
that are not captured by a simple block-diagonal structure.

\begin{figure}[htbp]
    \centering
    \includegraphics[width=0.9\linewidth]{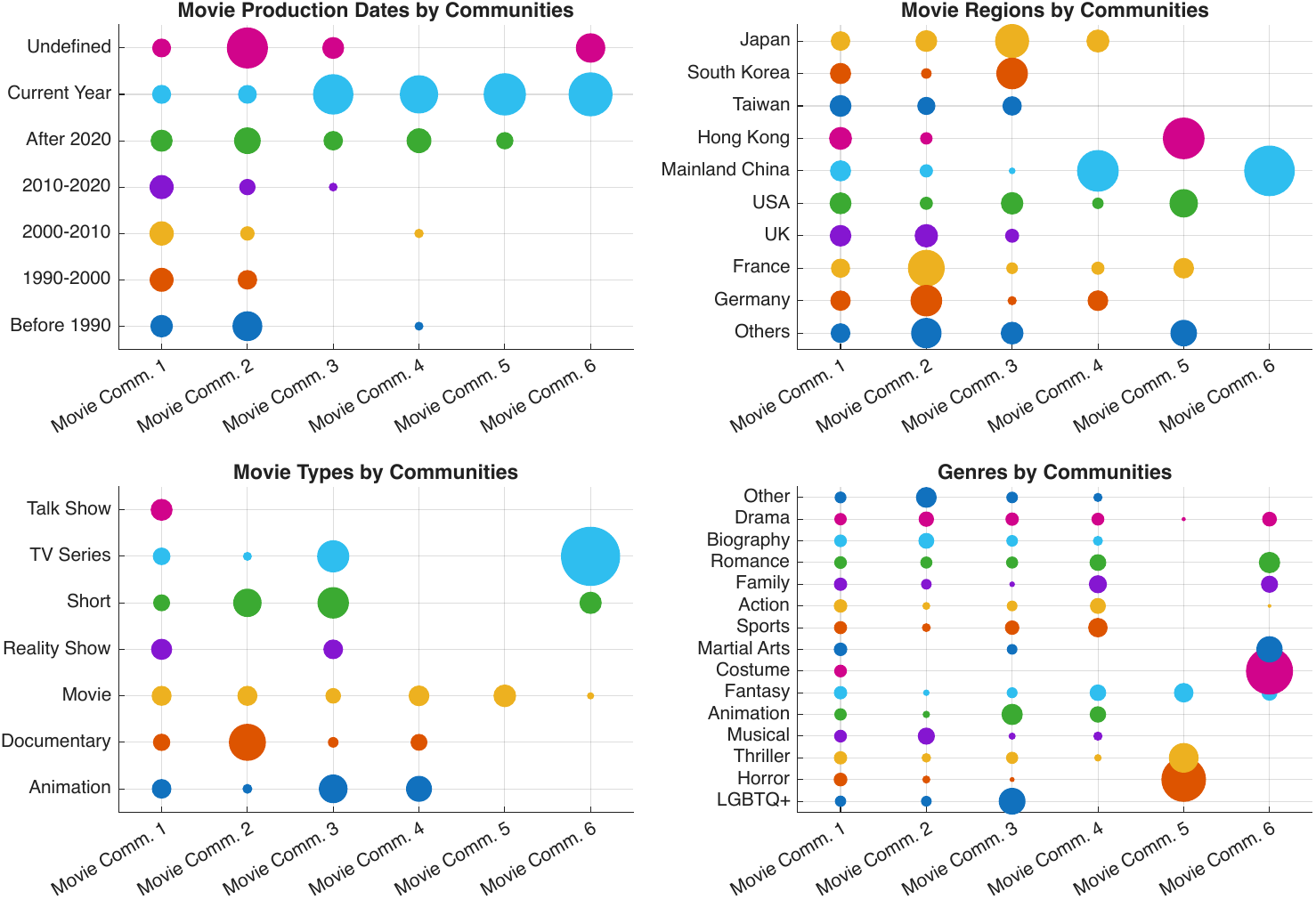}
    \caption{Analysis of movie communities using exogenous movie covariates: year of
    production, region, type, and genre. These covariates are not used in the clustering
    algorithm and are employed only for post-hoc interpretation.}
    \label{fig:movie}
\end{figure}

To further interpret the movie communities, we examine several auxiliary covariates that
were \emph{not} used in community detection: year of production, region, movie type, and
movie genre. Figure~\ref{fig:movie} summarizes the distributions of these covariates
within each movie community. The largest community, movie community~1, is broadly
representative and displays relatively balanced distributions across all covariates.
Movie community~2 consists mainly of older (pre-1990) European and UK productions, with a
high proportion of documentaries and musicals. In contrast, movie communities~3--6 are
dominated by more recent titles. Community~3 focuses on Japanese and South Korean TV
series and animations; Community~5 is enriched in horror and fantasy movies from Hong
Kong, the USA, and other regions. Communities~4 and~6 both correspond to mainland China
content, but with different profiles: Community~4 emphasizes mainstream movies, whereas
Community~6 is dominated by TV series, especially costume dramas. Combining these patterns
with Figure~\ref{fig:connectivity} gives a coherent interpretation of both user and movie communities.

This case study highlights several advantages of federated community detection. First,
although the five datasets have disjoint user sets and very different activity levels,
BiFLICKER yields a common set of user and movie communities. This avoids the
community matching problem that typically arises when clustering is performed separately
on each dataset. The global communities allow direct comparison of user populations across
datasets and provide a unified view of user preferences.

Second, federated learning naturally accommodates heterogeneity in user activity. In
Table~\ref{tab:comm}, the estimated communities are relatively balanced in dataset~1 but
highly unbalanced in datasets~4 and~5, where the majority of users belong to the
``general user'' community~1 due to their limited number of ratings. Nevertheless,
BiFLICKER still identifies meaningful representations of communities 2--6 within these
sparser datasets, borrowing strength from the denser datasets. Overall, the analysis
demonstrates that BiFLICKER can effectively uncover interpretable user and movie
communities in real-world bipartite networks under federated constraints.

\section{FLICKER: Convergence Analysis}\label{sec:flickerconv}
\subsection{Proof of Lemma 1}
In this section, we prove the motivating lemma in the main document.
\begin{proof}[Proof of Lemma~\ref{lem:opt}]
Let $A$ be a symmetric matrix with eigenvalues $\lambda_1 \geq \cdots \geq \lambda_n$
and orthonormal eigenvectors $\{\zeta_i\}_{i=1}^n$.
For any $v \in \mathbb{R}^n$ with $\|v\|=1$, we can write
$v = \sum_{i=1}^n a_i \zeta_i$ with $\sum_{i=1}^n a_i^2 = 1$.

Let $i^*$ be such that $|\lambda - \lambda_{i^*}| = \min_i |\lambda - \lambda_i|$.
Then the loss function
\[
\ell(v;\lambda)
= \frac{1}{2} \big\|(\lambda I - A)v\big\|^2
= \frac{1}{2} \sum_{i=1}^n (\lambda - \lambda_i)^2 a_i^2
\;\ge\; \frac{1}{2} (\lambda - \lambda_{i^*})^2 \sum_{i=1}^n a_i^2
= \frac{1}{2} (\lambda - \lambda_{i^*})^2,
\]
with equality if and only if $a_i = 0$ for all $i \neq i^*$, i.e.\ $v = \pm \zeta_{i^*}$.

For the Riemannian gradient of $\ell(v;\lambda)$ on the unit sphere
$S^{n-1} = \{v : \|v\|=1\}$, we have
\[
\nabla_{S^{n-1}} \ell(v;\lambda)
= (I - vv^\top)(\lambda I - A)^2 v.
\]
In particular, for each eigenvector $\zeta_i$ we have
$(\lambda I - A)^2 \zeta_i = (\lambda - \lambda_i)^2 \zeta_i$, and hence
\[
\nabla_{S^{n-1}} \ell(\zeta_i;\lambda)
= (I - \zeta_i \zeta_i^\top)(\lambda - \lambda_i)^2 \zeta_i = 0,
\qquad 1 \le i \le n.
\]
A direct calculation of the Riemannian Hessian shows that it is positive semidefinite at  $\zeta_{i^*}$.
\end{proof}

\subsection{Proof of Proposition 1}
In Proposition 1, we claim that using grid search will recover the full spectrum.

\begin{proof}
Under the run-to-convergence assumption, each estimate $\hat{\sigma}^{(i)}$ stored by the algorithm equals the exact singular
value associated with $v^{(i)}$. Consider an active interval with endpoints $\hat{\sigma}^{(i)}>\hat{\sigma}^{(i+1)}$, and let $\sigma$
be its squared-midpoint target value.

Suppose first that the interval contains an undiscovered singular value
$\sigma_0$. Since
$(\hat{\sigma}^{(i+1)})^2<\sigma_0^2<
(\hat{\sigma}^{(i)})^2$, we have
$|\sigma^2-\sigma_0^2|$ strictly smaller than the distance from
$\sigma^2$ to either endpoint. By assumption, the closest singular
value to $\sigma$ is therefore a unique singular value in the interior
of the interval. FLICKER recovers its singular vector, which is
orthogonal to both endpoint singular vectors. Hence the new pair is
inserted and the interval is subdivided.

If the interval contains no undiscovered singular value, its two
endpoints are equally closest to the squared target value, while all
other singular values are farther away. FLICKER therefore converges
to a unit vector in the span of the two endpoint singular vectors.
Such a vector has absolute inner product at least $1/\sqrt{2}$ with
one of them, and hence cannot satisfy both orthogonality conditions in Algorithm~\ref{alg:lambda}. The interval is therefore correctly
terminated.

Thus an interval is subdivided if and only if it contains an
undiscovered singular value, so all singular values are eventually
recovered. After the two initial calls, $\min\{m,n\}-2$ successful
calls discover the remaining singular values, and $\min\{m,n\}-1$ terminal calls certify the intervals between consecutive singular values. Hence the total number of FLICKER calls is
$2\min\{m,n\}-1$.
\end{proof}

\subsection{Proof of Theorem~1}
\begin{proof}[Proof of Theorem~\ref{thm:GD}]
Let $M=I-\eta(\lambda I-A)^2$ and
$\mu_i=1-\eta(\lambda-\lambda_i)^2$. Since each iteration of
Algorithm~\ref{alg:GD} applies $M$ followed by normalization, we have
$v^t=M^t v^0/\|M^t v^0\|$. Write
$v^0=\sum_{i=1}^n a_{0,i}u_i$. Since $Mu_i=\mu_i u_i$, it follows that
$M^t v^0=\sum_{i=1}^n \mu_i^t a_{0,i}u_i$, and hence
$a_{t,i}=\mu_i^t a_{0,i}/
\{\sum_{k=1}^n\mu_k^{2t}a_{0,k}^2\}^{1/2}$.
Therefore, for any $a_{0,j}\neq0$,
\[
\frac{a_{t,i}}{a_{t,j}}
=
\left(
\frac{1-\eta(\lambda-\lambda_i)^2}
     {1-\eta(\lambda-\lambda_j)^2}
\right)^t
\frac{a_{0,i}}{a_{0,j}}.
\]

Under the step-size condition, $0<\mu_i\le1$ for all $i$. Since $i^*$
is the unique minimizer of $|\lambda-\lambda_i|$, we have
$\mu_{i^*}>\mu_i$ for all $i\neq i^*$. Let
$r=\max_{i\neq i^*}\mu_i/\mu_{i^*}<1$. Using
$\sum_i a_{t,i}^2=1$ gives
\[
\frac{1-a_{t,i^*}^2}{a_{t,i^*}^2}
=
\sum_{i\neq i^*}
\left(\frac{\mu_i}{\mu_{i^*}}\right)^{2t}
\frac{a_{0,i}^2}{a_{0,i^*}^2}
\le
r^{2t}\frac{1-a_{0,i^*}^2}{a_{0,i^*}^2}.
\]
Thus
\[
1-a_{t,i^*}^2
\le
\frac{1-a_{0,i^*}^2}{a_{0,i^*}^2}r^{2t}.
\]

Let $s=\operatorname{sign}(a_{0,i^*})$. Since $\mu_{i^*}>0$, the sign
of $a_{t,i^*}$ does not change, and therefore
$\|v^t-su_{i^*}\|^2=2(1-|a_{t,i^*}|)
\le2(1-a_{t,i^*}^2)$. Consequently,
\[
\min_{s\in\{-1,1\}}\|v^t-su_{i^*}\|
\le
\frac{\sqrt{2}}{|a_{0,i^*}|}r^t.
\]

Finally, under the normalized Gaussian initialization, write
$v^0=g/\|g\|$ with $g\sim\mathcal N(0,I_n)$. Since
$\langle g,u_{i^*}\rangle\sim\mathcal N(0,1)$, a Gaussian small-ball
bound together with concentration of $\|g\|$ implies that, for any
$\delta\in(0,1)$, $|a_{0,i^*}|\ge c_0\delta/\sqrt n$ with probability
at least $1-\delta-\exp(-cn)$. Substituting this into the preceding
bound yields
\[
\min\{\|v^T-u_{i^*}\|, \|v^T + u_{i^*}\|\}
\le
\frac{C\sqrt n}{\delta}r^T.
\]
The stated iteration complexity follows.
\end{proof}

\section{Preparation}
\label{sec:aux-lemmas}

We collect several basic spectral inequalities used in the main proof.

\subsection{Spectral Bounds for Matrix Products}

\begin{lemma}[Spectral bounds for products]
\label{lem:condprod}
Let $A\in\reals^{n\times K}$ and $B\in\reals^{K\times p}$ with
$\mathrm{rank}(A)=\mathrm{rank}(B)=K$. Let $\sigma_1(\cdot)\ge \dots \ge \sigma_K(\cdot)$
denote singular values. Then
\[
\sigma_K(AB)\ge \sigma_K(A)\,\sigma_K(B),
\qquad
\sigma_1(AB)\le \sigma_1(A)\,\sigma_1(B).
\]
\end{lemma}

\begin{proof}
This lemma was introduced in \cite{hu2024network}. Here we prove it again.
Write the singular value decomposition $B = U D V^\top$ where
$U \in \reals^{K\times K}$, $V\in\reals^{p\times K}$ are orthonormal and
$D = \mathrm{diag}(\sigma_1(B),\dots,\sigma_K(B))$.
Then
\[
AB = (A U) D V^\top.
\]
By definition of singular values,
\[
\sigma_1(AB)
= \sqrt{\lambda_1\big((AB)(AB)^\top\big)}
= \sqrt{\lambda_1\big(AU D^2 U^\top A^\top\big)}.
\]
Using $D^2\preceq\sigma_1(B)^2I_K$ and congruence by $AU$, followed by eigenvalue monotonicity,
\[
\lambda_1(AU D^2 U^\top A^\top)
\le \lambda_1(D^2)\,\lambda_1( A U U^\top A^\top)
= \sigma_1(B)^2\,\lambda_1(AA^\top)
= \sigma_1(B)^2\,\sigma_1(A)^2.
\]
Taking square roots gives $\sigma_1(AB)\le \sigma_1(A)\sigma_1(B)$.

Similarly,
\[
\sigma_K(AB)
= \sqrt{\lambda_K\big(AU D^2 U^\top A^\top\big)}.
\]
Likewise, $D^2\succeq\sigma_K(B)^2I_K$ gives $AUD^2U^\top A^\top\succeq\sigma_K(B)^2AA^\top$, so
\[
\lambda_K(AU D^2 U^\top A^\top)
\ge \lambda_K(D^2)\,\lambda_K(AU U^\top A^\top)
= \sigma_K(B)^2\,\lambda_K(AA^\top)
= \sigma_K(B)^2\,\sigma_K(A)^2,
\]
and the lower bound for $\sigma_K(AB)$ follows.
\end{proof}

\subsection{\blue{Spectral properties of population matrix}}

\begin{lemma}
\label{lem:eigenOmega}
Under Assumption \ref{aspt:Bipart},
there exists a constant $C_1>0$, depending only on $\rho,L_1,L_2$, such that deterministically,
\begin{itemize}
\item[(i)] $C_1 \sqrt{nm}\theta_0 \ge \sigma_1(Q)\ge \sigma_L(Q)\ge \sqrt{nm}\theta_0/C_1$.
\item[(ii)] For each $i\in[n]$,
\[
{\|u_i\|\le\frac{C_1}{\sqrt{n}},}
\]
where $u_i$ is the $i$th row of $U$ (the left singular matrix of $Q$).
\item[(iii)] If $z_1(i)=z_1(j)$, then $u_i=u_j$. If $z_1(i)\neq z_1(j)$, then $\|u_i-u_j\|\ge \frac{1}{C_1\sqrt{n}}$.
\end{itemize}
Here $z_1:[n]\to[L_1]$ is the community label mapping induced by $Z_1$.
The same row-norm upper bound and community separation hold for $V$, with $m$ and $z_2$ in place of $n$ and $z_1$.
\end{lemma}

\begin{proof}
Let $\Omega=QQ\ic=Z_1PZ_1\ic$, where $P=GZ_2\ic Z_2G\ic$.
Its leading eigenvectors are the left singular vectors $U$ of $Q$.
The diagonal entries of $D_1=Z_1^\top Z_1$ and $D_2=Z_2^\top Z_2$
are the actual community sizes, at least $\rho n$ and $\rho m$, respectively.
Thus the nonzero singular values of $Z_1$ and $Z_2$ are of order $\sqrt n$
and $\sqrt m$. Also, $\sigma_L(G)\ge\rho\theta_0$ and
$\|G\|\le\sqrt{L_1L_2}\theta_0\le\theta_0/\rho$.
Applying Lemma~\ref{lem:condprod} to the factors obtained from a thin SVD of $G$, and using the community-size bounds, gives for some $C_0$,
\[
C_0^{-2} \theta_0^2 nm\leq \lambda_L(\Omega)\leq \lambda_1(\Omega)\leq C_0^2 \theta_0^2 nm.
\]
This proves (i).

The columns of $Z_1$ are indicators of disjoint sets, hence orthogonal with
$\|Z_1e_{l_1}\|^2=n_{l_1}$; thus $Z_1^\top Z_1=D_1$ and $\tilde Z_1:=Z_1D_1^{-1/2}$
satisfies $\tilde Z_1^\top\tilde Z_1=I_{L_1}$. Likewise
$\tilde Z_2:=Z_2D_2^{-1/2}$ has orthonormal columns. Since $Z_h=\tilde Z_hD_h^{1/2}$,
\[
Q=Z_1GZ_2^\top
=\tilde Z_1\big(D_1^{1/2}GD_2^{1/2}\big)\tilde Z_2^\top.
\]
The two outer matrices have orthonormal columns, so this factorization also identifies the nonzero singular values of $Q$.

Next, let $Q=U\Lambda V\ic$ be the SVD, we find that
\[
U =Q V \Lambda^{-1}=Z_1 R,\quad R=
GZ_2\ic V \Lambda^{-1}.
\]
Note that if $z_1(i)=z_1(j)$, then $e_i\ic Z_1=e_j\ic Z_1$, so $e_i\ic U=e_j\ic U$. Likewise, we can write the right singular vectors as $V=Z_2 S$ for $S=G\ic Z_1\ic U \Lambda^{-1}$.

For a left community $l_1$, set $u^\top=R(l_1,{:})$ and $n_{l_1}=|\{i:z_1(i)=l_1\}|$. Since $U=Z_1R$, $U(i,{:})=u^\top$ for every $i$ with $z_1(i)=l_1$.
Thus $n_{l_1}uu^\top=\sum_{i:z_1(i)=l_1}U(i,{:})^\top U(i,{:})\preceq U^\top U=I_L$, giving $\|u\|\le n_{l_1}^{-1/2}\le(\rho n)^{-1/2}$; the same argument applies to $V$.
For $z_1(i)=l_1\ne l_1'=z_1(j)$, condition (iii) gives $l_2$ with $|G(l_1,l_2)-G(l_1',l_2)|\ge\rho\theta_0$. Since $Q(i,r)=G(z_1(i),z_2(r))$,
\[
\begin{aligned}
\|Q(i,{:})-Q(j,{:})\|^2
&=\sum_{r=1}^m\{G(l_1,z_2(r))-G(l_1',z_2(r))\}^2\\
&\ge |\{r:z_2(r)=l_2\}|\,\{G(l_1,l_2)-G(l_1',l_2)\}^2
\ge\rho^3m\theta_0^2.
\end{aligned}
\]
Hence $\|Q(i,{:})-Q(j,{:})\|\ge\rho^{3/2}\theta_0\sqrt m$.
On the other hand, $Q=U\Lambda V^\top$ implies
$\|Q(i,{:})-Q(j,{:})\|\le\sigma_1(Q)\|u_i-u_j\|$.
Combining with (i) proves (iii). Transposing $Q$ proves the right-side conclusions.
\end{proof}

\section{\texorpdfstring{$\ell_2$}{l2} and \texorpdfstring{$\ell_{2,\infty}$}{l2,infty} Perturbation}
\label{sec:l2-l2inf}

We now state and prove an $\ell_2$ and $\ell_{2,\infty}$ perturbation bound for singular vectors, specialized to the symmetric case for clarity. It is used in Step 2 of the recovery proof.
Let $\twoninf{A} := \max_i \|e_i^\top A\|_2$ denote the row-wise $\ell_2$ sup-norm.

\begin{proposition}[$\ell_2$ and $\ell_{2,\infty}$ perturbation]
\label{prop:linfty}
Let $A,\Omega \in \reals^{n\times n}$ be symmetric matrices, and suppose
$\Omega$ has rank $L$ with eigendecomposition
$\Omega = U \Lambda U^\top$, where $U\in\reals^{n\times L}$ has orthonormal columns and
$\Lambda = \mathrm{diag}(\lambda_1,\dots,\lambda_L)$ with
$\lambda_1\ge \dots \ge \lambda_L>0$. Assume a spectral gap
\[
\lambda_L - \lambda_{L+1} \ge n G_1 \theta_0^2
\]
for some $G_1>0$, with the convention $\lambda_{L+1}=0$.
Let $\widehat{U}\in\reals^{n\times L}$ be the matrix of leading $L$ orthonormal eigenvectors of $A$.

Suppose there exists $b_0>0$ such that
\[
\|A-\Omega\| \le b_0 \theta_0^2 n,\qquad
\|A-\Omega\|_{2,\infty}\le C_0 \theta_0\sqrt{(n+m)\log n/m},
\qquad
\|(A-\Omega)U\|_{2,\infty} \le b_0 \theta_0^2 \sqrt{n},
\]
with $b_0$ small enough (e.g.\ $b_0\le G_1/22$). Assume also $\|U\|_{2,\infty}\le C_U/\sqrt n$ for a fixed $C_U$; the constants below may depend on $C_U$.
Then there exist constants $C_1,C_2>0$ and an orthogonal matrix
$O\in\reals^{L\times L}$ such that
\[
\|U O- \widehat{U} \| \le C_1 \frac{b_0}{G_1},
\qquad
\|U O- \widehat{U} \|_{2,\infty} \le C_2 \frac{b_0}{G_1 \sqrt{n}}(1+\frac{\sqrt{(n+m)\log n}}{G_1\theta_0\sqrt{nm}}).
\]
\end{proposition}

\begin{proof}

Set $E = A-\Omega$. Let $U_\perp\in\reals^{n\times (n-L)}$ be an orthonormal basis of
the orthogonal complement of $\mathrm{span}(U)$, so that $[U,U_\perp]$ is orthogonal.
We can expand $\widehat{U}$ as
\[
\widehat{U} = U R + \Delta_\perp,
\]
where $R = U^\top \widehat{U} \in \reals^{L\times L}$ and
$\Delta_\perp = U_\perp U_\perp^\top \widehat{U}$ satisfies $U^\top \Delta_\perp = 0$.

{\bf Step 1: Spectral norm control.}
By the Davis--Kahan theorem \citep{daviskahan} applied to $A$ and $\Omega$,
the spectral gap condition and $\|E\|\le b_0 n\theta_0^2$ imply
\[
\|\Delta_\perp\| \le C \frac{\|E\|}{n G_1 \theta_0^2}
\le C \frac{b_0}{G_1},
\]
for some universal constant $C>0$. In particular, taking $C\le 2\sqrt{2}$, we can assume
$\|\Delta_\perp\|\le 2\sqrt{2}\,b_0/G_1$. Since
\[
\widehat{U}^\top \widehat{U}
= (U R + \Delta_\perp)^\top (U R + \Delta_\perp)
= R^\top R + \Delta_\perp^\top \Delta_\perp
= I_L,
\]
we have
\[
\|R^\top R - I_L\|
= \|\Delta_\perp^\top \Delta_\perp\|
\le \|\Delta_\perp\|^2
\le 8 b_0^2 / G_1^2.
\]
Thus all singular values of $R$ lie in $[1- 8b_0^2/G_1^2, 1+8b_0^2/G_1^2]$.
Let $R = \Psi L \Phi^\top$ be the SVD of $R$, with $L$ diagonal and orthogonal
$\Psi,\Phi\in\reals^{L\times L}$. Define the orthogonal matrix $O := \Psi \Phi^\top$.
Then
\[
\|R - O\|
= \|\Psi L \Phi^\top - \Psi I \Phi^\top\|
= \|L - I\|
\le 8 b_0^2 / G_1^2.
\]

We now bound the spectral norm difference between $U$ and $\widehat{U}O$:
\begin{align*}
\|\widehat{U} - U O\|
&\le \|U(R - O)\| + \|\Delta_\perp\|\\
&\le \|R - O\| + \|\Delta_\perp\|
\le 8 b_0^2/G_1^2 + 2\sqrt{2}\,b_0/G_1.
\end{align*}
For $b_0\le G_1/22$ we can absorb the quadratic term into the linear one and obtain
\[
\|\widehat{U} - U O\|
\le C_1 \frac{b_0}{G_1}
\]
for some constant $C_1$.

{\bf Step 2: Row-wise $\ell_2$ norm control.}
We next bound $\twoninf{\widehat{U} - U O}$.
Let $e_j^\top$ denote the $j$th standard basis row. Then
\[
e_j^\top(\widehat{U} - U O)
= e_j^\top \Delta_\perp + e_j^\top U (R - O).
\]
The second term is easily controlled by
\[
\|e_j^\top U (R - O)\|
\le \|e_j^\top U\|\,\|R - O\|
\le \|e_j^\top U\|\, 8 b_0^2/G_1^2.
\]
The row bound gives $\|e_j^\top U\|\le C_U/\sqrt n$, so this term is
$O(b_0^2/(G_1^2\sqrt{n}))$.

It remains to bound $\|e_j^\top \Delta_\perp\|$ uniformly over $j$. Using the eigenspace
equation for $\widehat{U}$ and the fact that $U^\top \Delta_\perp = 0$, one can show
(by applying $I-UU^\top$ to $A\widehat U=\widehat U\widehat\Lambda$) that
\[
\Delta_\perp = (I - UU^\top) E \widehat{U} \widehat{\Lambda}^{-1},
\]
where $\widehat{\Lambda}$ is the diagonal matrix of top $L$ eigenvalues of $A$.
Therefore,
\[
e_j^\top \Delta_\perp
= e_j^\top (I - UU^\top) E \widehat{U} \widehat{\Lambda}^{-1}.
\]
Expanding $\widehat{U} = U R + \Delta_\perp$ and using $\Omega U = U\Lambda$ and
$(I-UU^\top)\Omega = 0$ yields
\[
e_j^\top \Delta_\perp
= e_j^\top (I-UU^\top)E(UR + \Delta_\perp)\widehat{\Lambda}^{-1}.
\]
So we can bound
\begin{align*}
\|e_j^\top \Delta_\perp\|
&\leq \|e_j^\top E UR \widehat{\Lambda}^{-1}\|
+\|e_j\ic E \Delta_\perp\widehat{\Lambda}^{-1}\|+
\|e_j^\top UU^\top E \widehat{U}\widehat{\Lambda}^{-1}\|
\\
&\leq \|EU\|_{2,\infty}\|R \widehat{\Lambda}^{-1}\|
+\|e_j\ic E\| \|\Delta_\perp\|\|\widehat{\Lambda}^{-1}\|
+\|e_j^\top U\|\|U^\top E\| \|\widehat{U}\|\|\widehat{\Lambda}^{-1}\|
\end{align*}
Using the assumptions
$\|(A-\Omega)U\|_{2,\infty} \le b_0\theta_0^2\sqrt{n}$ and $\|A-\Omega\|\le b_0\theta_0^2 n$, together with the
eigenvalue gap (which implies $\|\widehat{\Lambda}^{-1}\|\lesssim (\theta_0^2 n G_1)^{-1}$),
$\|\Delta_\perp\|\lesssim b_0/G_1$ and the row-norm upper bound on $U$, we arrive at
\begin{align*}
\|e_j^\top \Delta_\perp\|
&\leq \|EU\|_{2,\infty}\|R \widehat{\Lambda}^{-1}\|
+\|e_j\ic E\| \|\Delta_\perp\|\|\widehat{\Lambda}^{-1}\|
+\|e_j^\top U\|\|U^\top E\| \|\widehat{U}\|\|\widehat{\Lambda}^{-1}\|\\
&\lesssim \frac{b_0}{G_1\sqrt{n}}+\frac{b_0C_0\sqrt{(n+m)\log n}}{G^2_1 n \theta_0\sqrt{m}}
+\frac{C_U b_0}{G_1\sqrt n}
\lesssim \frac{b_0}{G_1\sqrt{n}}(1+\frac{\sqrt{(n+m)\log n}}{G_1\theta_0\sqrt{mn}}).
\end{align*}
So
\[
\|e_j^\top \Delta_\perp\|
\le C'\,\frac{b_0}{G_1\sqrt{n}}(1+\frac{\sqrt{(n+m)\log n}}{G_1\theta_0\sqrt{nm}})
\]
for a constant $C'$ depending only on $G_1$ and $C_0$-type bounds. Combining the two
components,
\[
\|e_j^\top(\widehat{U} - U O)\|
\le \|e_j^\top \Delta_\perp\| + \|e_j^\top U (R - O)\|
\le C_2 \frac{b_0}{G_1\sqrt{n}}(1+\frac{\sqrt{(n+m)\log n}}{G_1\theta_0\sqrt{nm}}),
\]
for some $C_2>0$, uniformly over $j$. This yields
\[
\|\widehat{U} - U O\|_{2,\infty}
\le C_2 \frac{b_0}{G_1\sqrt{n}}(1+\frac{\sqrt{(n+m)\log n}}{G_1\theta_0\sqrt{nm}}),
\]
as claimed.
\end{proof}
\blue{
\begin{lemma}
\label{lem:l2inf-E}
Let $A=\tfrac1m BB^\top$ and $\Omega=\tfrac1m QQ^\top$, and set $W:=B-Q$.
Under Assumption~\ref{aspt:Bipart}, if $\min\{n,m\}\theta_0\ge c_0\log(n+m)$ for a sufficiently
large constant $c_0>0$, then there is a constant $C>0$ (depending only on $\rho,L_1,L_2$) such that, with probability at least $1-n^{-2}$,
\[
\twoninf{A-\Omega}
=\max_{i\in[n]}\frac1m\big\|e_i^\top(BB^\top-QQ^\top)\big\|_2
\;\le\; C\,\theta_0\sqrt{(n+m)\log n/m}.
\]
\end{lemma}
}
\begin{proof}
Since $B=Q+W$ with $\E W=0$,
\begin{equation}\label{eq:gram-decomp}
BB^\top-QQ^\top = QW^\top + WQ^\top + WW^\top .
\end{equation}
Writing $q_i^\top=e_i^\top Q$ and $w_i^\top=e_i^\top W$ for the $i$-th rows and using
$e_i^\top QW^\top=(Wq_i)^\top$, $e_i^\top WQ^\top=(Qw_i)^\top$,
$e_i^\top WW^\top=(Ww_i)^\top$, the triangle inequality gives
\begin{equation}\label{eq:three-terms}
\big\|e_i^\top(BB^\top-QQ^\top)\big\|_2
\le \|W q_i\|_2+\|Q w_i\|_2+\|W w_i\|_2 .
\end{equation}

\emph{Preliminary bounds.} With probability at least $1-n^{-2}$, uniformly in $i\in[n]$:
\begin{itemize}
\item[(P1)] $\|q_i\|_2\le\sqrt{m}\,\theta_0$, since $\|q_i\|_2^2=\sum_j Q_{ij}^2\le m\theta_0^2$
      (deterministic, as $Q_{ij}\le\theta_0$).
\item[(P2)] $\|W\|\le C_W\sqrt{(n+m)\theta_0}$, by the operator-norm
      bound obtained from Corollary 3.12 and Remark 3.13 of \citet{bandeira2016sharp} applied to the symmetric dilation of $W$. (The variance term is at most $(n+m)\theta_0$ and the entry bound is one)
\item[(P3)] $\|w_i\|_2\le\sqrt{3m\theta_0}$. Indeed $B_{ij}^2=B_{ij}$ gives
      $\|w_i\|_2^2=\sum_j(B_{ij}-Q_{ij})^2\le\sum_j B_{ij}+\sum_j Q_{ij}^2=d_i+\|q_i\|_2^2$,
      where $d_i=\sum_j B_{ij}$ is the degree. Since $\E d_i\le m\theta_0$, a Chernoff bound
      and a union bound over $i$ (using $m\theta_0\ge c_0\log n$) yield
      $\max_i d_i\le 2m\theta_0$, so $\|w_i\|_2^2\le 2m\theta_0+m\theta_0^2\le 3m\theta_0$.
\item[(P4)] $\|Z_2^\top w_i\|_2\le C_Z\sqrt{m\theta_0\log n}$. Each coordinate
      $(Z_2^\top w_i)_\ell=\sum_{j:z_2(j)=\ell}W_{ij}$ is a sum of independent mean-zero
      terms with $|W_{ij}|\le1$ and $\sum_j Var(W_{ij})\le N_{2,\ell}\theta_0\le m\theta_0$.
      Bernstein's inequality with $t\asymp\sqrt{m\theta_0\log n}\le m\theta_0$ gives
      $\Prob(|(Z_2^\top w_i)_\ell|\ge t)\le n^{-3}$; a union bound over $i\in[n]$ and
      $\ell\in[L_2]$ and $L_2=O(1)$ give the claim.
\end{itemize}

\emph{Term 1 ($QW^\top$-part $\|Wq_i\|$).} By (P2) and (P1),
\[
\|W q_i\|_2\le\|W\|\,\|q_i\|_2\le C_W'\sqrt{(n+m)\theta_0}\cdot\sqrt{m}\,\theta_0
= C_W'\, \sqrt{(n+m)m}\,\theta_0^{3/2}.
\]

\emph{Term 2 ($WQ^\top$-part $\|Qw_i\|$).} Since
$\|Z_1\|=\sqrt{\max_{l_1} N_{1,l_1}}\le\sqrt n$ and $\|G\|\le\theta_0/\rho$ by
Assumption~\ref{aspt:Bipart}(ii) and $L_1,L_2\le1/\rho$,
\[
\begin{aligned}
\|Q w_i\|_2&=\|Z_1 G (Z_2^\top w_i)\|_2
\le\|Z_1\|\,\|G\|\,\|Z_2^\top w_i\|_2
\\
&\le \frac{C_Z\theta_0}{\rho}\sqrt{nm\theta_0\log n}
= \frac{C_Z}{\rho}\sqrt{nm}\,\theta_0^{3/2}\sqrt{\log n},
\end{aligned}
\]
using (P4).

\emph{Term 3 ($WW^\top$-part $\|Ww_i\|$).} By (P2) and (P3),
\[
\|W w_i\|_2\le\|W\|\,\|w_i\|_2\le C_W'\sqrt{(n+m)\theta_0}\cdot\sqrt{3m\theta_0}
= \sqrt3\,C_W'\, \sqrt{m(n+m)}\,\theta_0 .
\]

\emph{Combine.} Insert the three bounds into \eqref{eq:three-terms}, divide by $m$:
\[
\begin{aligned}
\frac1m\big\|e_i^\top(BB^\top-QQ^\top)\big\|_2
&\le C_W'\theta_0^{3/2}\sqrt{\frac{n+m}{m}}
+\frac{C_Z}{\rho}\theta_0^{3/2}\sqrt{\frac{n\log n}{m}}
+\sqrt3 C_W'\theta_0\sqrt{\frac{n+m}{m}}\\
&\le C\big(\theta_0+\theta_0^{3/2}\sqrt{\log n}\big)\sqrt{(n+m)/m}.
\end{aligned}
\]
Taking the maximum over $i$ (all bounds hold uniformly) proves the claim; the final
inequality uses $\theta_0\le1$.
\end{proof}

\section{Consistency of BiFLICKER}
\label{sec:strong-consistency}

In this section we prove the weak and strong consistency conclusions of Theorem~\ref{thm:sequential-community-recovery}, using local spectral calibration, Proposition~\ref{prop:linfty}, and a $k$-means label recovery lemma.

Let $\widehat{U}$ and $\widehat{V}$ denote the matrices of leading $L$ left and right
singular vectors of $B$, respectively.
Let $\widehat{U}^T$ and $\widehat{V}^T$ denote the corresponding approximations after
$T$ iterations of the federated power iteration \gdpi\ used in BiFLICKER, initialized with independent Gaussian vectors and using only the reference-block singular values as targets, as in the sequential version of Algorithm~\ref{alg:bi}. The initialization and projections are specified in Step 3.

\subsection{Proof of Theorem~\ref{thm:sequential-community-recovery}}

We give the statistical calculations for the left side and explicitly control the right-side and numerical errors below. In the theoretical clustering argument, $k$-means means global minimization of its objective.

{\bf Step 1: Local vs global singular values.}
Let $B_k\in\{0,1\}^{n_k\times m}$ denote the adjacency matrix on server $k$, so that the
rows correspond to a subset $\mathcal{L}^{(k)}\subset\mathcal{L}$ and the columns to the
full set $\mathcal{R}$. Let $Q_k = \E[B_k\mid\mathcal L^{(k)}] = Z_1^{(k)} G Z_2^\top$ denote the local mean
matrix, where $Z_1^{(k)}$ is the submatrix of $Z_1$ with rows restricted to
$\mathcal{L}^{(k)}$.

Write
\[
Q_k^\top Q_k
= Z_2 G^\top (Z_1^{(k)})^\top Z_1^{(k)} G Z_2^\top
= Z_2 G^\top D^{(k)} G Z_2^\top,
\]
where $D^{(k)} = (Z_1^{(k)})^\top Z_1^{(k)}$ is diagonal with entries equal to the
community sizes in $\mathcal{L}^{(k)}$.
We first verify Proposition~\ref{prop:local-gram-calibration} of the main paper for a uniformly
sampled reference block, with sampling independent of the edge variables.
Let $D_1=Z_1^\top Z_1$ contain the actual global community sizes.
For each $l_1$, $D^{(k)}(l_1,l_1)$ is a sum of $n_k$ draws without replacement from $\{\mathbf1\{z_1(i)=l_1\}:i\in[n]\}$, whose mean is $D_1(l_1,l_1)/n$ and variance is at most one. Bernstein's inequality for sampling without replacement (\href{https://arxiv.org/pdf/1309.4029}{Bardenet and Maillard, 2015, Proposition 1.4}) gives
\[
\Pr\!\left(\left|D^{(k)}(l_1,l_1)-\frac{n_k}{n}D_1(l_1,l_1)\right|\ge t\right)
\le2\exp\!\left\{-\frac{t^2}{2(n_k+t/3)}\right\}.
\]
Taking $t=C\sqrt{n_k}\log n$ and a union bound over $l_1$, we obtain $\|(n/n_k)D^{(k)}-D_1\|\le Cn\log n/\sqrt{n_k}\le C\sqrt n\log n$ with probability at least $1-2L_1e^{-c\log^2n}$, using $n_k\ge\rho n$. The exact identity
\[
\frac n{n_k}Q_k^\top Q_k-Q^\top Q
=Z_2G^\top\left(\frac n{n_k}D^{(k)}-D_1\right)GZ_2^\top
\]
gives $\|(n/n_k)Q_k^\top Q_k-Q^\top Q\|\le C\theta_0^2m\sqrt n\log n$, since $\|Z_2\|\le\sqrt m$ and $\|G\|\le C\theta_0$.
On the good event in \eqref{eqn:B_Q} of Step 2, $\|E_B\|\le C_2\sqrt{(n+m)\theta_0}$, where $E_B=B-Q$.
Lemma~\ref{lem:eigenOmega}(i) gives $\|Q\|\le C\theta_0\sqrt{nm}$, so
\[
\begin{aligned}
B^\top B-Q^\top Q&=Q^\top E_B+E_B^\top Q+E_B^\top E_B,\\
\|B^\top B-Q^\top Q\|
&\le2\|Q\|\|E_B\|+\|E_B\|^2
\le C\{\theta_0^{3/2}\sqrt{nm(n+m)}+\theta_0(n+m)\}.
\end{aligned}
\]
For $E_{B_k}=B_k-Q_k$, row restriction gives $\|E_{B_k}\|\le\|E_B\|$ and $\|Q_k\|\le\|Q\|$. Thus the scaled local error satisfies $(n/n_k)\|B_k^\top B_k-Q_k^\top Q_k\|\le\rho^{-1}\{2\|Q\|\|E_B\|+\|E_B\|^2\}$.
Since $(n+m)/(nm\theta_0)=1/(n\theta_0)+1/(m\theta_0)\le C/\log n$, we have $\theta_0(n+m)\le C\theta_0^{3/2}\sqrt{nm(n+m)}/\sqrt{\log n}$. Combining these estimates yields
\[
\left\|\frac n{n_k}B_k^\top B_k-B^\top B\right\|
\le C\{\theta_0^2m\sqrt n\log n+\theta_0^{3/2}\sqrt{nm(n+m)}\}.
\]
Weyl's inequality then gives
\begin{equation}
\label{eq:local-global-sigma}
\left|
\frac{n}{n_k} \sigma_i(B_k)^2 - \sigma_i(B)^2
\right|
= O(\sqrt{n}\log n\, m \theta_0^2+\theta_0^{3/2}\sqrt{nm(n+m)}),
\end{equation}
with high probability. This justifies using
$\widetilde\sigma_i = \sqrt{n/n_k}\,\sigma_i(B_k)$ as an approximate tuning value
for the target singular value in \gdpi.

{\bf Step 2: Spectral perturbation for $U$ and $\widehat{U}$.}
Consider the symmetric matrix
\[
A := \frac{1}{m} BB^\top,\qquad
\Omega := \frac{1}{m} QQ^\top.
\]
Note that $\Omega$ has eigen-decomposition
$\Omega = U \Lambda U^\top$ with
$\Lambda = m^{-1} \Sigma^2$ and the same left singular vectors $U$ as $Q$.

\blue{
Using Bernstein inequality it is easy to derive that with high probability $\|B\|_\infty$ and $\|B\|_1$ which are the max degree of nodes, are bounded by $C_2 m\theta_0$ and $C_2 n\theta_0$ for some $C_2$ with high probability.
For $E_B:=B-Q$, with high probability,
\be\label{eqn:B_Q}
\begin{aligned}
\|E_B\| &= \left\|\left(\begin{array}{ll}
0 &E_B\\
E_B\ic &0
\end{array}\right)\right\|\\
&\leq C\{\sqrt{(n+m)\theta_0}+\sqrt{\log(n+m)}\}
\leq C_2\sqrt{(n+m)\theta_0}.
\end{aligned}
\ee
Here we apply Corollary 3.12 and Remark 3.13 of \citet{bandeira2016sharp} to the symmetric dilation, as in Lemma~\ref{lem:l2inf-E}. All noise bounds in this step can be taken on a joint event of probability at least $1-C(n+m)^{-2}$: use $\log(n+m)$ in the tail bounds, and use $\log(n+m)=O(\log n)$ to retain the displayed rates. Since $m\le n^\alpha$, the conditions $n\theta_0/\log n$ and $m\theta_0/\log n$ sufficiently large imply $\min(n,m)\theta_0\gtrsim\log(n+m)$, which absorbs the logarithmic term. Recall that $\Omega = \frac{1}{m}QQ\ic$. This leads to
\be\label{eqn:BB_QQ}
\begin{aligned}
\frac{1}{m}\|BB\ic-QQ\ic\|
&\leq \frac{2\|Q\|\|E_B\|+\|E_B\|^2}{m}\\
&\leq C\left\{\theta_0^{3/2}\sqrt{\frac{n(n+m)}{m}}+\frac{\theta_0(n+m)}{m}\right\}
\leq C_3\theta_0\sqrt{\theta_0 n(n+m)/m}.
\end{aligned}
\ee
for some $C_3$.  Since $\lambda_L(\Omega)\ge nG_1\theta_0^2$ for a fixed $G_1>0$ and $\lambda_{L+1}(\Omega)=0$, Lemma~\ref{lem:eigenOmega} verifies the signal-to-zero gap in Proposition~\ref{prop:linfty}. No internal signal gap is needed.}

Davis--Kahan gives
$\min_O\|U-\widehat UO\|_F\le C\sqrt{(n+m)/(\theta_0nm)}$,
since $L$ is bounded. Thus the centralized embedding has the required
weak-consistency rate; Step 3 will include the finite-iteration error.
Next we consider the $2\to \infty$ norm, note
\begin{equation}
\label{tmp:BBdecomp}
\frac{1}{m}\|(BB\ic-QQ\ic)U\|_{2,\infty}
\leq \frac{1}{m}\|(B-Q)Q\ic U\|_{2,\infty}+
\frac{1}{m}\|B(B-Q)\ic U\|_{2,\infty}
\end{equation}
We check the first term. Write $Q = U \Lambda_Q V\ic$ as the SVD, then $\|\Lambda_Q\| = \|Q\| \leq C \sqrt{nm} \theta_0$. For $i$-th row,
\[
\frac{1}{m}\|e_i\ic(B-Q)Q\ic U\|=\frac{1}{m}\|e_i\ic(B-Q)V\Lambda_Q\|\leq C\sqrt{n/m}\theta_0 \|e_i\ic(B-Q)V\|.
\]
For fixed $i$ and $l\le L$, the summands $E_B(i,r)V(r,l)$ are independent and centered. Lemma~\ref{lem:eigenOmega}(ii) gives $|E_B(i,r)V(r,l)|\le C_2/\sqrt m$, and $\sum_r\operatorname{var}(E_B(i,r)V(r,l))\le\theta_0\sum_rV(r,l)^2=\theta_0$. Bernstein's inequality therefore gives
\[
\Pr\!\left(\left|\sum_{r=1}^mE_B(i,r)V(r,l)\right|\ge t\right)
\le2\exp\!\left\{-\frac{t^2}{2(\theta_0+C_2t/(3\sqrt m))}\right\}.
\]
Take $t=C\{\sqrt{\theta_0\log(n+m)}+\log(n+m)/\sqrt m\}$. A union bound over $i,l$ gives $\max_i\|e_i^\top E_BV\|\le\sqrt L\,t$ except with probability $C(n+m)^{-2}$.

The conditions $m\theta_0\gtrsim\log(n+m)$ and $\log(n+m)=O(\log n)$ imply $\sqrt L\,t\le C\sqrt{\theta_0\log n}$.
In conclusion, $\frac1m\|(B-Q)Q\ic U\|_{2,\infty}\leq C_3\theta_0^{3/2}\sqrt{n\log n/m}
$ with high probability under our condition.

For the second term in \eqref{tmp:BBdecomp}, write $E:=E_B=B-Q$ and split
$B=Q+E$, so that
\begin{equation}\label{eqn:split2}
B(B-Q)\ic U = QE\ic U + EE\ic U .
\end{equation}

\emph{The $QE\ic U$ part.} Let $q_i\ic=e_i\ic Q$ denote the $i$th row of $Q$.
Since $Q_{ij}\le\theta_0$ we have $\|q_i\|\le\theta_0\sqrt m$ deterministically, so by
\eqref{eqn:B_Q},
\[
\frac1m\big\|e_i\ic QE\ic U\big\|
=\frac1m\big\|q_i\ic E\ic U\big\|
\le\frac1m\|q_i\|\,\|E\|
\le\frac{1}{m}\,\theta_0\sqrt m\cdot C_2\sqrt{(n+m)\theta_0}
= C\,\theta_0^{3/2}\sqrt{\tfrac{n+m}{m}} .
\]
\emph{The $EE\ic U$ part.}  Writing $E_{i\cdot}\in\reals^{1\times m}$ for the
$i$th row of $E$ and $u_k\ic=e_k\ic U$ for the $k$th row of $U$,
\[
e_i\ic EE\ic U
=\sum_{k=1}^n\langle E_{i\cdot},E_{k\cdot}\rangle u_k\ic
=\underbrace{\|E_{i\cdot}\|^2\,u_i\ic}_{=:T_1}
+\underbrace{E_{i\cdot}M^{(i)}}_{=:T_2},
\qquad
M^{(i)}:=\sum_{k\neq i}E_{k\cdot}\ic u_k\ic\in\reals^{m\times L},
\]
where $M^{(i)}$ is independent of $E_{i\cdot}$ because the rows of $E$ are independent.

For $T_1$, the bound $\|E_{i\cdot}\|^2\le 3m\theta_0$ from the proof of
Lemma~\ref{lem:l2inf-E} together with $\|u_i\|\le C_1/\sqrt n$ from
Lemma~\ref{lem:eigenOmega}(ii) gives $\frac1m\|T_1\|\le C\theta_0/\sqrt n$.

For $T_2$, each entry of $M^{(i)}$ is $M^{(i)}_{j\ell}=\sum_{k\neq i}E_{kj}U_{k\ell}$, a
sum of independent mean-zero terms with
$\sum_k\text{var}(E_{kj}U_{k\ell})\le\theta_0\sum_kU_{k\ell}^2=\theta_0$ and
$|E_{kj}U_{k\ell}|\le C_1/\sqrt n$. Bernstein's inequality and a union bound over
$j\in[m]$, $\ell\in[L]$ give, with probability at least $1-n^{-10}$,
\[
\max_{j\in[m]}\|M^{(i)}_{j\cdot}\|\le C\sqrt{\theta_0\log n},
\]
using $n\theta_0\ge\log n$. Conditionally on $M^{(i)}$, the quantity
$T_2=\sum_{j=1}^mE_{ij}M^{(i)}_{j\cdot}$ is a sum of $m$ independent mean-zero vectors
with $\sum_j\text{var}(E_{ij}M^{(i)}_{j\ell})\le\theta_0\sum_j(M^{(i)}_{j\ell})^2\le
Cm\theta_0^2\log n$ and $|E_{ij}M^{(i)}_{j\ell}|\le C\sqrt{\theta_0\log n}$, so a second
application of Bernstein's inequality yields
\[
\frac1m\|T_2\|
\le C\Big(\frac{\theta_0\log n}{\sqrt m}+\frac{\sqrt{\theta_0}\,(\log n)^{3/2}}{m}\Big)
\le C\,\frac{\theta_0\log n}{\sqrt m},
\]
where the last step uses $m\theta_0\ge\log n$.

Combining the two parts of \eqref{eqn:split2}, and using
$\theta_0/\sqrt n+\theta_0\log n/\sqrt m
\le C\theta_0^{3/2}\sqrt{(n+m)\log n/m}$
since $n\theta_0\gtrsim\log n$ and $(n+m)\theta_0\gtrsim\log n$, we conclude
\[
\frac{1}{m}\|(BB\ic-QQ\ic)U\|_{2,\infty}\leq C\theta_0^{3/2}\sqrt{(n+m)\log n/m}.
\]

Summarizing, we have the following by using Lemma \ref{lem:l2inf-E}:
\[
\|A-\Omega\|
\lesssim b_0\theta_0^2 n,
\qquad
\|A-\Omega\|_{2,\infty}\lesssim \theta_0 \sqrt{(m+n)\log n/m},\quad
\|(A-\Omega)U\|_{2,\infty}
\lesssim b_0 \blue{\theta_0^2}\sqrt{n},
\]
where we can take
\[
b_0 \asymp \sqrt{\frac{(n+m)\log n}{nm\theta_0}}.
\]
\blue{Note that Lemma \ref{lem:eigenOmega} has verified all conditions on the population matrix for Proposition~\ref{prop:linfty}.}
For exact recovery, now impose $\min(n,m)\theta_0^2/\log n$ sufficiently large. Since $b_0^2\lesssim\theta_0\{\log n/(n\theta_0^2)+\log n/(m\theta_0^2)\}$ and $\theta_0\le1$, this ensures the smallness condition of Proposition~\ref{prop:linfty}.
Applying Proposition~\ref{prop:linfty}, we obtain an orthogonal matrix
$O_L\in\reals^{L\times L}$ such that
\[
\begin{aligned}
\|U - \widehat{U} O_L\|
&=O(b_0),\\
\|U - \widehat{U} O_L\|_{2,\infty}
&= O\left(\frac{b_0}{\sqrt{n}}(1+\frac{\sqrt{(n+m)\log n}}{G_1\theta_0\sqrt{nm}})\right),
\end{aligned}
\]
with high probability.
The additional condition gives $(n+m)\log n/(nm\theta_0^2)=\log n/(n\theta_0^2)+\log n/(m\theta_0^2)\le C$, so the left row bound is $Cb_0/\sqrt n$.
Transposing the preceding calculations, with $A=B^\top B/n$ and
$\Omega=Q^\top Q/n$, yields on the same event
$\min_O\|V-\widehat VO\|_F\le C\sqrt{(n+m)/(nm\theta_0)}$
and, under the stronger condition, $\min_O\|V-\widehat VO\|_{2,\infty}\le Cb_0/\sqrt m$.
Taking the constant in the additional condition sufficiently large gives $\sqrt n\min_O\|U-\widehat UO\|_{2,\infty}\le Cb_0\le c_1/2$ and $\sqrt m\min_O\|V-\widehat VO\|_{2,\infty}\le Cb_0\le c_1/2$, where $c_1$ is the constant in Lemma~\ref{lem:simplelabel}.
The two sides may use different orthogonal rotations.

{\bf Step 3: Federated iteration error.}
By Lemma \ref{lem:eigenOmega}, $\sigma_L(Q)^2\asymp nm\theta_0^2$.
The convergence analysis of $\gdpi$ in Theorem \ref{thm:GD} (when spectrum is simple) and Theorem \ref{thm:GDproj} (when spectrum is degenerate) uses the spectral weights $1-\eta(\lambda-\lambda_i)^2$. We apply its matrix-power argument to the complete signal subspace as follows.

For this calculation, set $A=BB^\top/s$, $\widetilde\lambda_j=\widetilde\sigma_j^2/s$, and $M_j=I-s^2\eta_j(A-\widetilde\lambda_jI)^2$, where $s=Cnm\theta_0^2$ with fixed large $C$. We are interested in the spectral gap of $M_j$ over different eigenvectors.

Steps 1--2 give $0\preceq A\preceq I$, $\lambda_L(A)-\lambda_{L+1}(A)\ge c_0>0$, and $\max_{j\le L}|\widetilde\lambda_j-\lambda_j(A)|\le C\{\log n/\sqrt n+\sqrt{(n+m)/(nm\theta_0)}\}\le C\{\log n/\sqrt n+1/\sqrt{\log n}\}=o(1)$; the step-size condition gives $c_1I\preceq M_j\preceq I$ and $s^2\eta_j\asymp1$.
Write $\lambda_1(A)\ge\cdots\ge\lambda_n(A)$. Among the $L$ intervals $(c_0(8L)^{-h},c_0(8L)^{-(h-1)}]$, $1\le h\le L$, one contains none of the $L-1$ internal gaps. Let $\tau$ be its lower endpoint and define $\Pi_j:=\sum_{i\in\mathcal J_j}\widehat U({:},i)\widehat U({:},i)^\top$, and
\[
\begin{aligned}
\mathcal J_j&=\left\{i\in[L]:\lambda_r(A)-\lambda_{r+1}(A)\le\tau
\text{ for every }\min(i,j)\le r<\max(i,j)\right\},
\end{aligned}
\]
where $\widehat U({:},i)$ is the $i$th leading unit eigenvector of $A$.
Each $\mathcal J_j$ consists of consecutive indices and contains $j$. By telescoping, its spectral width satisfies
$\max_{i\in\mathcal J_j}\lambda_i(A)-\min_{i\in\mathcal J_j}\lambda_i(A)=\sum_{r=\min\mathcal J_j}^{\max\mathcal J_j-1}\{\lambda_r(A)-\lambda_{r+1}(A)\}\le(|\mathcal J_j|-1)\tau\le(L-1)\tau$.
For $i\in\mathcal J_j$ and $i'\in[L]\setminus\mathcal J_j$, an intervening gap exceeds $\tau$ and none lies in $(\tau,8L\tau]$, so $|\lambda_i(A)-\lambda_{i'}(A)|>8L\tau$. For $i\in\mathcal J_j$ and $i'>L$, $|\lambda_i(A)-\lambda_{i'}(A)|\ge\lambda_L(A)-\lambda_{L+1}(A)\ge c_0\ge8L\tau$.
The corresponding eigenvalues of $M_j$ are $\mu_{j,i}=1-s^2\eta_j\{\lambda_i(A)-\widetilde\lambda_j\}^2$, $i\in[n]$. In summary,
$|\lambda_i(A)-\widetilde\lambda_j|\le(L-1)\tau+o(1)\le L\tau$ for $i\in\mathcal J_j$, and $|\lambda_i(A)-\widetilde\lambda_j|\ge8L\tau-o(1)$ otherwise. Hence
\[
\min_{i\in\mathcal J_j}\mu_{j,i}-\max_{i\in[n]\setminus\mathcal J_j}\mu_{j,i}
\ge s^2\eta_j\{(8L\tau-o(1))^2-(L\tau)^2\}\ge\kappa>0.
\]

Let $v_j=v_j^{(T)}$ be the $j$th left singular-vector estimate returned by sequential BiFLICKER, Algorithm~\ref{alg:bi}, after $T$ projected FLICKER updates with target $\widetilde\sigma_j^2$, so $\widehat U^T=[v_1,\ldots,v_L]$ before orthogonal alignment.
Set $V_0=[v_1,\ldots,v_{j-1}]$, $P_j=I-V_0V_0^\top$, and $\varepsilon_{j-1}=\max_{i<j}\|(I-\Pi_i)v_i\|$, with $\varepsilon_0=0$. Set $W=[\Pi_1v_1,\ldots,\Pi_{j-1}v_{j-1}]$. Orthogonality gives $\|W-V_0\|\le\sqrt L\varepsilon_{j-1}$ and $\|W^\top W-I\|\le L\varepsilon_{j-1}^2$: within each group $(W^\top W-I)(i,i')=-v_i^\top(I-\Pi_i)v_{i'}$, and all other entries vanish. Thus, for small $\varepsilon_{j-1}$, $V_*=W(W^\top W)^{-1/2}$ and $P_*=I-V_*V_*^\top$ satisfy $\|P_j-P_*\|\le C\varepsilon_{j-1}$ (at $j=1$, use $P_*=I$). Let $v_{*,i}=V_*({:},i)$. The matrix $W^\top W$ is block diagonal by spectral group, so $v_{*,i}\in\operatorname{range}(\Pi_i)$ and the $v_{*,i}$ are orthonormal. Consequently,
\[
\begin{aligned}
\Pi_jV_*V_*^\top
&=V_*V_*^\top\Pi_j=\sum_{i<j:\Pi_i=\Pi_j}v_{*,i}v_{*,i}^\top,\\
S_j:=P_*\Pi_j
&=\Pi_j-\sum_{i<j:\Pi_i=\Pi_j}v_{*,i}v_{*,i}^\top.
\end{aligned}
\]
Thus $S_j^\top=S_j^2=S_j$, and its rank is
\[
\begin{aligned}
d_j:=\operatorname{rank}(S_j)=\operatorname{tr}(S_j)
&=|\mathcal J_j|-|\mathcal J_j\cap\{1,\ldots,j-1\}|\\
&=|\mathcal J_j\cap\{j,\ldots,L\}|\ge1,
\end{aligned}
\]
because $j\in\mathcal J_j$.
The matrix $P_*M_jP_*$ preserves its range, and its eigenvalues there exceed those on the complement by at least $\kappa$. For $H_j=P_jM_jP_j$, Weyl and Davis--Kahan yield
\[
\begin{gathered}
 \|H_j-P_*M_jP_*\|\le C\varepsilon_{j-1},\qquad
 \|\widehat S_j-S_j\|\le C\varepsilon_{j-1},\qquad
 \frac{\lambda_{d_j+1}(H_j)}{\lambda_{d_j}(H_j)}\le1-\kappa/2=:r<1,
\end{gathered}
\]
where $\widehat S_j$ projects onto the leading $d_j$ eigenvectors of $H_j$.

For independent $g_j\sim N(0,I_n)$, the first update is $z_j=P_jM_jg_j/\|P_jM_jg_j\|$ and $v_j=H_j^{T-1}z_j/\|H_j^{T-1}z_j\|$. All denominators are nonzero almost surely since $M_j\succeq c_1I$. Conditional on the preceding stages and the data, choose unit $h\in\operatorname{range}(\widehat S_j)$. Since $h^\top M_jg_j\sim N(0,\|M_jh\|^2)$ and $\|M_jh\|\ge c_1$, Gaussian bounds give $\|\widehat S_jz_j\|\ge |h^\top M_jg_j|/\|g_j\|\ge c_1(n+m)^{-3}$ except with probability $C(n+m)^{-2}$. Consequently,
\[
\begin{aligned}
 \|(I-\widehat S_j)v_j\|
 &\le\frac{r^{T-1}}{\|\widehat S_jz_j\|}\le C(n+m)^3r^T,\\
 \varepsilon_j&\le C\varepsilon_{j-1}+C(n+m)^3r^T
 \le C_L(n+m)^3r^T,\\
 \min_O\|[v_1,\ldots,v_L]-\widehat UO\|_F
 &\le\sqrt2\|(I-\widehat U\widehat U^\top)[v_1,\ldots,v_L]\|_F
 \le\sqrt{2L}\varepsilon_L.
\end{aligned}
\]
The second line follows by induction from $\varepsilon_0=0$ with fixed $L$; the inequality uses $\|(I-\Pi_j)v_j\|\le\|(I-\widehat S_j)v_j\|+\|\widehat S_j-S_j\|$. Taking $T=\lceil C_T\log(n+m)\rceil$ with sufficiently large $C_T$ makes $C_L(n+m)^3r^T\le(n+m)^{-2}$, verifies the smallness conditions at every stage, and gives total failure probability at most $CL(n+m)^{-2}$.

{\bf Step 4: Federated $k$-means clustering.}
Thus, for a step size $\eta\asymp (n^2 m^2 \theta_0^4)^{-1}$ and
$T=\lceil C_T\log(n+m)\rceil$ iterations, the eigenvector iterates produced by the centralized gradient
iteration based on the approximate spectral parameter $\widetilde\sigma_i^2$ satisfy
\[
\min_O\|\widehat{U}^T - \widehat{U}O\|_F \le \frac{1}{n+m}
\]
with high probability. Orthogonal rotations preserve $k$-means distances, so we align the computed embedding and retain the notation $\widehat U^T$. In particular,
\[
\|\widehat{U}^T - \widehat{U}\|_{2,\infty}
\le \|\widehat{U}^T - \widehat{U}\|_F
\le \frac{1}{n+m}.
\]
Combining this with the bound in Step~2 by the triangle inequality yields
\[
\|U - \widehat{U}^T O_L\|_{2,\infty}
\le \|U - \widehat{U} O_L\|_{2,\infty}
    + \|\widehat{U} - \widehat{U}^T\|_{2,\infty}
= O\Big(b_0/\sqrt{n}\Big).
\]
This row bound uses the additional exact-recovery condition in Step 2. Including the numerical errors, the scaled row errors on both sides are at most $c_1/2+1/\sqrt{\min(n,m)}\le c_1$ for large $n,m$. Under the basic conditions, the Frobenius bound instead gives $\min_O\|\widehat U^T-UO\|_F^2\le C(n+m)/(nm\theta_0)$.

We now invoke a $k$-means label recovery Lemma \ref{lem:simplelabel}, stated and proved in the next
subsection, to convert the $\ell_{2,\infty}$ bound into exact cluster recovery. The
assumption of well-separated population centers for the rows of $U$ ensures that, for
$n$ large enough and the error bound above small enough, every global minimizer of the $k$-means objective gives the true partition, up to a permutation of labels.
Thus BiFLICKER recovers $z_1$ exactly with high probability. The argument for $V$ and
$z_2$ is identical. In the right iteration, $A=B^\top B/s$ has the same nonzero spectrum and we use independent $g_j\sim N(0,I_m)$. The Frobenius bounds and the weak-recovery part of the same lemma give the stated misclassification rates under the basic conditions.
\hfill$\square$

\subsection{Label Recovery via \texorpdfstring{$k$-means}{k-means}}
\label{subsec:kmeans-proof}

We restate and prove the $k$-means label recovery lemma used in the previous step.

\begin{lemma}[$k$-means label recovery]
\label{lem:simplelabel}
Let $U\in\reals^{n\times L}$ have population rows $u_i$, and suppose
there exist centers $m_1,\dots,m_{L_1}\in\reals^L$ such that
\[
u_i = m_{z_1(i)},\qquad i=1,\dots,n,
\]
and the centers are well separated:
\[
\|m_{l_1}-m_{l_1'}\|\ge\frac\gamma{\sqrt n},\qquad l_1\ne l_1',
\]
for some $\gamma>0$. Assume every community contains at least $\rho n$ nodes. \blue{Let $\widehat{U}\in\reals^{n\times L}$ be an estimator of $U$. Let $\widehat z_1$ globally minimize the $k$-means objective on its rows with $L_1$ centers. There exist sufficiently small constants $c_1>0$ and $q_0\in(0,1)$ such that, for $0<q\le q_0$, if there is an orthogonal matrix $O$ with
\[
\|U O-\Uhat\|_F\leq c_1q,
\]
then there is a permutation, $\pi$, so that
\[
\frac{1}{n}|\{i:  \hat z_1(i)\neq\pi(z_1(i)) \}|\leq q^2.
\]
If in addition,} we have
\[
\max_{i\le n} \|u_i O - \widehat{u}_i\|
\le \frac{c_1}{\sqrt{n}},
\]
with the same sufficiently small $c_1$.
Then, for $n$ large enough and $c_1$ sufficiently
small, there exists a permutation $\pi$ of $\{1,\dots,L_1\}$ such that
\[
\widehat{z}_1(i) = \pi(z_1(i)),\qquad i=1,\dots,n.
\]
\end{lemma}
\begin{proof}
Write $\ell=z_1$ and let $\hat\ell=\widehat z_1$ be the minimizing labels. We use row vectors throughout, so the aligned centers are $m_{l_1}O$.
Define the index set
\[
I := \bigl\{i\in[n]: \|u_i O - \hat{u}_i\|\le \frac{c_1}{\sqrt{n}}\bigr\}.
\]
For every $i\notin I$, $\|u_i O - \hat{u}_i\|^2 > c_1^2/n$, so
\[
|I^c|\cdot \frac{c_1^2}{n}
\le \sum_{i\in I^c} \|u_i O - \hat{u}_i\|^2
\le \|U O - \Uhat\|_F^2
\le c_1^2 q^2,
\]
which implies $|I^c|\le q^2 n$ and hence $|I|\ge n(1-q^2)$.
In the second scenario where the row-wise norm is bounded by $\max_i \|u_i O - \hat{u}_i\|\le c_1/\sqrt{n}$, we have $I=[n]$.

Consider $k$-means on $\{\hat{u}_i\}$. For any labels $z:[n]\to[L_1]$ and centers
$m_1,\dots,m_{L_1}$, define
\[
L(z,m) := \sum_{i=1}^n n\|\hat{u}_i - m_{z(i)}\|^2.
\]
Let $(\hat\ell,\hat m)$ be a minimizer of $L(z,m)$ over all choices of $(z,m)$, and
consider the population configuration $(\ell,m)$.

First note that
\[
L(\ell,m O)
= \sum_{i=1}^n n\|\hat{u}_i - m_{\ell(i)} O\|^2
= \sum_{i=1}^n n\|\hat{u}_i - u_i O\|^2
\le nc_1^2q^2
\]
By optimality of $(\hat\ell,\hat m)$,
\begin{equation}
\label{eq:tmp-Lupp}
L(\hat\ell,\hat m)\le L(\ell,m O)\le c_1^2 q^2n.
\end{equation}

Next fix a community $l_1$ and denote $Q_{l_1}:=\{i:\ell(i)=l_1\}$, $n_{l_1}:=|Q_{l_1}|\ge \rho n$.
Define
\[
\pi(l_1) := \arg\min_{1\le j\le L_1} \|m_{l_1} O- \hat m_j\|.
\]
Then for any $i\in Q_{l_1}$,
by $\|u+v\|^2\ge \tfrac12\|v\|^2 - \|u\|^2$,
\[
\|\hat{u}_i - \hat m_{\hat\ell(i)}\|^2
\ge \frac12\|m_{l_1} O - \hat m_{\hat\ell(i)}\|^2 - \|\hat{u}_i-m_{l_1}O\|^2.
\]
Summing over $i\in Q_{l_1}$,
\[
\sum_{i\in Q_{l_1}}\|\hat{u}_i - \hat m_{\hat\ell(i)}\|^2
\ge \frac12\sum_{i\in Q_{l_1}}\|m_{l_1} O - \hat m_{\hat\ell(i)}\|^2
- \sum_{i\in Q_{l_1}}\|\hat{u}_i - m_{l_1} O\|^2.
\]
Using
$\|m_{l_1} O - \hat m_{\pi(l_1)}\|^2
\le \frac{1}{n_{l_1}} \sum_{i\in Q_{l_1}}\|m_{l_1} O- \hat m_{\hat\ell(i)}\|^2$,
we obtain
\[
\sum_{i\in Q_{l_1}}\|\hat{u}_i - \hat m_{\hat\ell(i)}\|^2
\ge \frac12 n_{l_1} \|m_{l_1} O - \hat m_{\pi(l_1)}\|^2 -\sum_{i\in Q_{l_1}}\|\hat u_i-m_{l_1}O\|^2.
\]
Summing over $l_1$ and combining with \eqref{eq:tmp-Lupp},
\[
L(\hat\ell,\hat m)
\ge \frac12\sum_{l_1=1}^{L_1} n_{l_1} n\|m_{l_1} O- \hat m_{\pi(l_1)}\|^2 -L(\ell,mO),
\]
so
\[
\frac12\sum_{l_1=1}^{L_1} nn_{l_1} \|m_{l_1} O - \hat m_{\pi(l_1)}\|^2
\le L(\hat\ell,\hat m) + L(\ell,m O)
\le 2c_1^2 q^2 n .
\]
Using $n_{l_1}\ge \rho n$ for all $l_1$, we can derive the upper bound for $\|m_{l_1} O - \hat{m}_{\pi(l_1)}\|^2$, and hence
\[
\|m_{l_1} O- \hat m_{\pi(l_1)}\|
\le \frac{2c_1 q}{\sqrt{\rho n}}.
\]
Therefore, recall that the centers are $\gamma/\sqrt{n}$-distance apart. If the distance
$\gamma/\sqrt n>4c_1q/\sqrt{\rho n}$,  two distinct population centers cannot share their matched estimated center. Hence $\pi$ is a permutation, and each matched pair satisfies
\[
\|\hat m_{\pi(l_1)} - m_{l_1} O\|
\le \frac{2c_1 q}{\sqrt{\rho n}}.
\]

Now fix $i\in I$ and let $l_1=\ell(i)$ denote the true community label. Suppose the estimated label $\hat\ell(i)\neq \pi(l_1)$.
Define an alternative label vector $\hat\ell'$ by changing only the $i$-th label:
$\hat\ell'(i)=\pi(l_1)$ and $\hat\ell'(j)=\hat\ell(j)$ for $j\neq i$, with centers $\hat m$
unchanged. Then
\[
L(\hat\ell',\hat m) - L(\hat\ell,\hat m)
= n\|\hat{u}_i - \hat m_{\hat\ell'(i)}\|^2 - n\|\hat{u}_i - \hat m_{\hat\ell(i)}\|^2.
\]
We bound these two terms. First,
\[
\|\hat{u}_i - \hat m_{\hat\ell'(i)}\|
= \|\hat{u}_i - \hat m_{\pi(l_1)}\|
\le \|\hat{u}_i - m_{l_1} O\| + \|m_{l_1} O- \hat m_{\pi(l_1)}\|
\le 2c_1/\sqrt{n} + \frac{2c_1 q}{\sqrt{n\rho}}.
\]
Using the assumptions on $c_1$ and $q$, this can be bounded by $4c_1/\sqrt{\rho n}$, so
\[
\|\hat{u}_i - \hat m_{\hat\ell'(i)}\|\le \frac{4c_1}{\sqrt{n\rho}}.
\]

On the other hand, write $\hat\ell(i)=\pi(l_1')$ with $l_1'\ne l_1$.
The distance to this wrong center is at least
$(\gamma-c_1-2c_1q/\sqrt\rho)/\sqrt n$, whereas the distance to the
matched center is at most $(c_1+2c_1q/\sqrt\rho)/\sqrt n$.
Choose $c_1,q_0$ so that $2c_1+4c_1q_0/\sqrt\rho<\gamma$.
The wrong-center distance is then strictly larger.
Thus $L(\hat\ell',\hat m) < L(\hat\ell,\hat m)$, contradicting the optimality of
$(\hat\ell,\hat m)$. Therefore, for all $i\in I$, we must have $\hat\ell(i)=\pi(\ell(i))$.

For weak consistency, since $|I^c|\le q^2 n$, misclassifications can only occur in $I^c$, and hence
\[
\frac{1}{n}|\{i:\hat\ell(i)\neq \pi(\ell(i))\}|
\le \frac{|I^c|}{n}\le q^2.
\]

Finally, under the additional rowwise assumption
$\max_i \|u_i O - \hat{u}_i\|\le c_1/\sqrt{n}$, we have $I=[n]$, so
$\hat\ell(i)=\pi(\ell(i))$ for all $i$, i.e. strong consistency.

\end{proof}

\clearpage
\bibliography{ref_clean}
\end{document}